\pdfoutput=1

\documentclass[11pt]{article}

\usepackage{acl}

\usepackage{times}
\usepackage{latexsym}
\usepackage[T1]{fontenc}
\usepackage[utf8]{inputenc}
\usepackage{microtype}
\usepackage{inconsolata}
\usepackage{xcolor}

\usepackage{latexsym}
\usepackage{adjustbox}
\usepackage{xspace}
\usepackage{todonotes}
\usepackage{xcolor}
\usepackage{comment}
\usepackage{tikz}
\usepackage{pgfplots}
\usepackage{pgfplotstable}
\usepackage{makecell}
\usepackage{adjustbox}
\usepackage{pifont}
\usepackage{float}
\usepackage{supertabular}
\usepackage{multirow}
\usepackage{longtable}
\usepackage{graphicx}
\usepackage[edges]{forest}
\definecolor{hidden-draw}{RGB}{0,0,0}
\usepackage{wrapfig}
\usepackage{CJKutf8}
\usepackage{CJK}

\usepackage{tcolorbox}
\tcbset{
    mybox/.style={
        fontupper=\linespread{.8}\selectfont,
        sharp corners,
        top=0pt, 
        bottom=0pt 
    }
}

\usepackage{amsmath}
\usepackage{amssymb}

\usepackage{booktabs}
\usepackage{threeparttable}
\usepackage{tabularx}

\usepackage{soul}

\usepackage{subfig}

\usepackage{bm} 

\usepackage{algorithm}
\usepackage{algorithmic}

\usepackage{setspace}

\newcommand{\ctext}[2]{%
    \tikz[baseline=(X.base)] \node[fill=#1,rounded corners=2.8pt,inner sep=1pt] (X) {#2};%
}

\definecolor{mygreen}{RGB}{11,141,10}
\definecolor{myred}{RGB}{223,68,52}
\definecolor{myblue}{RGB}{70,130,180}
\definecolor{mydeepblue}{RGB}{65,105,225}
\definecolor{myviolet}{RGB}{97,0,138}
\definecolor{myburgundy}{RGB}{110,10,30}
\definecolor{myblue2}{RGB}{0,105,148}
\definecolor{iceblue}{RGB}{173, 216, 230}
\definecolor{puregreen}{RGB}{0, 218, 0}
\definecolor{graygreen}{RGB}{74,113,106}

\definecolor{wingreen}{rgb}{0,0.45,0.24}
\definecolor{losered}{rgb}{1.0,0.1,0.24}

\definecolor{lightcoral}{rgb}{0.97, 0.36, 0.46}
\definecolor{lightyellow}{rgb}{0.98, 0.7, 0}
\definecolor{harvestgold}{rgb}{0.85, 0.57, 0.0}
\definecolor{brightlavender}{rgb}{0.75, 0.58, 0.89}
\definecolor{capri}{rgb}{0.0, 0.75, 1.0}
\definecolor{carminepink}{rgb}{0.92, 0.3, 0.26}
\definecolor{celadon}{rgb}{0.67, 0.88, 0.69}
\definecolor{darkpastelgreen}{rgb}{0.01, 0.75, 0.24}

\definecolor{grayhighlight}{RGB}{250,250,227}

\definecolor{target}{HTML}{F47983}
\definecolor{control}{HTML}{3E87CD}
\definecolor{credibility}{HTML}{B98AC9}
\definecolor{logical}{HTML}{93C572}
\definecolor{emotional}{HTML}{F9EAC3}

\newenvironment{packeditemize}{
\begin{list}{$\bullet$}{
\setlength{\labelwidth}{8pt}
\setlength{\itemsep}{0pt}
\setlength{\leftmargin}{\labelwidth}
\addtolength{\leftmargin}{\labelsep}
\setlength{\parindent}{0pt}
\setlength{\listparindent}{\parindent}
\setlength{\parsep}{0pt}
\setlength{\topsep}{3pt}}}{\end{list}}

\usepackage{pifont}

\newcommand{\diffdown}[1]{\raisebox{0.5pt}{\fontsize{6}{5.5}\selectfont{\textcolor{wingreen}{\textbf{$\blacktriangledown${$ #1$}}}}}}

\newcommand{\placeholder}[1]{\ctext{yellow!40}{\{#1\}}}

\newcommand{\usepalatino}[1]{{\fontfamily{ppl}\selectfont #1}}

\newcommand{\warning}{\raisebox{2pt}{\fontencoding{U}\fontfamily{futs}\selectfont\char 49\relax}}

\newcommand{\myline}{\par
  \kern0pt 
  \hrule height 0.6pt
  \kern3pt 
}

\newcommand{\mylinenoskip}{\par
  \kern3pt 
  \hrule height 0.6pt
  \kern3pt 
}

\newcommand{\translate}[1]{\textcolor{gray}{#1}}

\newcommand{\cpt}{\texttt{PoC} tokens\xspace}
\newcommand{\detector}{\textsc{PocDetect}\xspace}
\newcommand{\trace}{\textsc{PocTrace}\xspace}

\newcommand{\poc}{\texttt{PoC}\xspace}

\usepackage{amsthm}

\usepackage{url}
\title{Speculating LLMs' Chinese Training Data Pollution from Their Tokens\\
\small
\usepalatino{\textcolor{orange}{\warning{}Caution: this paper may include offensive and upsetting content.}}
}

\author{Qingjie Zhang$^{1}$, Di Wang$^{1}$, Haoting Qian$^{1}$, Yan Liu$^{2}$, Tianwei Zhang$^{3}$, \\ \bf   Ke Xu$^{1,4}$, Qi Li$^{1,4}$, Minlie Huang$^{1,4}$, Hewu Li$^{1,4}$, Han Qiu$^{1,4*}$ \\
$^{1}$Tsinghua University, $^{2}$Ant Group, \\ $^{3}$Nanyang Technological University, $^{4}$Zhongguancun Laboratory\\
\texttt{Emails: qj-zhang24@mails.tsinghua.edu.cn, qiuhan@tsinghua.edu.cn}
}

\begin{document}
\begin{CJK}{UTF8}{gbsn}
\maketitle

\def\thefootnote{*}\footnotetext{Corresponding author.}\def\thefootnote{\arabic{footnote}}

\begin{abstract}

Tokens are basic elements in the datasets for LLM training. 
It is well-known that many tokens representing Chinese phrases in the vocabulary of GPT (4o/4o-mini/o1/o3/4.5/4.1/o4-mini)\footnote{Same for GPT-5 and -oss (August 26, 2025).} are indicating contents like pornography or online gambling. 
Based on this observation, \\
\textit{our goal is to locate \underline{Po}lluted \underline{C}hinese (\poc) tokens in LLMs and study the relationship between \poc tokens' existence and training data}.
(1) We give a formal definition and taxonomy \hfill of \poc tokens based on the GPT's vocabulary. \\
(2) We build a \poc token detector via fine-tuning an LLM to label \poc tokens in vocabularies by considering each token's both semantics \hfill and related contents from the search engines. \\
(3) We study how to speculate training data pollution via \poc tokens' appearances (token ID). 
Experiments on GPT and other 23 LLMs indicate that \poc tokens widely exist while GPT's vocabulary behaves the worst:  more than 40\% long Chinese tokens (i.e., a token with more than two Chinese characters) are either porn or online gambling. 
We validate our speculation method on famous pre-training datasets like C4 and Pile. 
Then, considering GPT-4o, we speculate the ratio of ``波*野结衣''\footnote{A Japanese porn star's name in Chinese (partially masked for privacy concerns) and also a token with index 185,946.} related webpages in its training data is around 0.5\%. 

\end{abstract}

\section{Introduction}
\label{sec:intro}

LLMs are pre-trained on enormous data crawled from the Internet. Consequently, polluted contents like pornography or online gambling are inevitably mixed into the crawled data. 
Without careful data cleaning, these contents may generate polluted tokens (or glitch tokens) when building vocabularies and performing tokenization like Byte-Pair Encoding (BPE) \cite{wang2020neural,sennrich2015neural}. 
One typical example is that there are various \underline{Po}lluted \underline{C}hinese tokens (\poc) in GPT-4o's vocabulary. 
Chinese native speakers can naturally realize that many of these \poc tokens (one token containing more than two Chinese characters) refer to illegal (i.e., porn or gambling) or anomalous contents\footnote{\url{https://gist.github.com/ctlllll}}. 
Later on, this vocabulary is incorporated by OpenAI to advanced GPT models including GPT-o1/o3/4.5/4.1/o4-mini~\cite{GPTtokens}. 

\begin{figure}[t]
    \centering
    \includegraphics[width=0.99\linewidth]{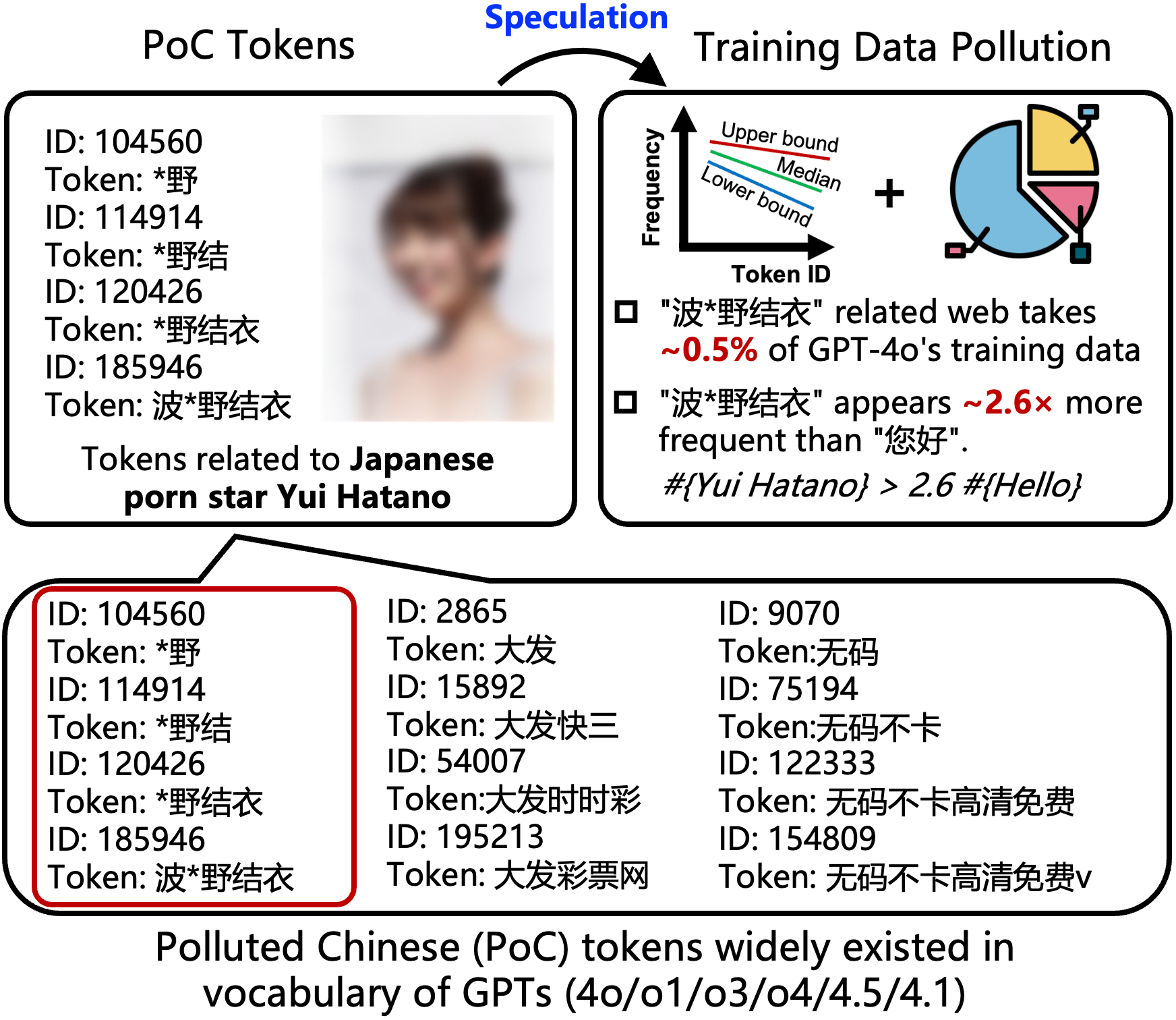}  
    \vspace{-2em}
    \caption{Overview: we aim to perform a systematic study on \poc tokens starting from GPT's vocabulary. Additionally, we try to address a challenging question: how to speculate GPT-4o's training data pollution from its vocabulary. Photo is blurred for privacy concerns.} 
    \label{fig:first}
    \vspace{-3ex}
\end{figure}

It is indicated that GPT-4o cannot explain some of its own \poc tokens\footnote{\url{https://github.com/openai/tiktoken/issues/297}}. 
Similar phenomena are also studied by previous works~\cite{li2024glitch,wang2024tokenization,land2024fishing}, which disclose that under-trained (or glitch) tokens can stimulate the LLM to generate inappropriate contents or have hallucinations. 
However, none of existing works give a rigorous study on these \poc tokens and investigate the relationship between their appearance and the training data pollution. 

In this paper, as shown in \autoref{fig:first}, we aim to \textit{conduct a systematic study on \poc tokens in contemporary LLMs and study how to speculate training data pollution by the \poc tokens in vocabularies}. 
Our insight is that the appearance of these \poc tokens indicates the pollution of the training dataset. 
Thus, based on a rigorous study of \poc tokens, we can speculate the polluted Chinese contents in both open-sourced large-scale training datasets and closed-sourced LLMs' training datasets like GPT-4o. 
Our research has three steps.

(1) \textbf{Expert labeling of GPT's \poc tokens.} 
There are 1,600+ long Chinese tokens with more than two Chinese characters in GPT's vocabulary. 
It is not easy to locate \poc tokens since a few Chinese characters are too implicit to understand. 
For instance, a GPT's token ``青青草'' (ID 56,167, translated as ``green grass'') refers to a famous pornographic software upon examination using a search engine. 
Relying on an expert team with sufficient knowledge about Chinese linguistics and culture, we give a formal definition and taxonomy to help experts label the GPT's tokens. 

(2) \textbf{Detecting \poc tokens in other LLMs.}
Based on the labeling of GPT's \poc tokens, we explore automatically locating \poc tokens in other LLMs. 
It is worth noting that Chinese LLMs like GLM have 28,000+ Chinese tokens, which are difficult for human labeling. 
Thus, we fine-tune an LLM to label Chinese tokens by combining their literal meanings and search engine results. 

(3) \textbf{Speculating training data from \poc tokens.} 
We further connect the token's appearance (i.e. token ID) to its frequency in the dataset. 
We give empirical estimation and verify on several famous open-sourced datasets. 
Then, based on this estimation method, we speculate the pollution ratio of GPT-4o's Chinese training data via some representative \poc tokens. 
Please kindly note that we are not OpenAI so there is no ground truth for GPT-4o's training data (\autoref{fig:first}). Still, we can verify this by poisoning open-sourced datasets to reproduce the appearance of GPT's \poc tokens. 

Our key findings are as follows. 
By detecting 9 vocabularies of 23 LLMs, we find that \poc tokens widely exist. 
By estimation and verification, we find that Chinese corpus in open-sourced datasets like mC4~\cite{xue2020mt5} is polluted: 2-3\% Chinese contents are polluted. 
In the end, by taking token ``波*野结衣'' and its three subsequence tokens (\autoref{fig:first}) as an example, we speculate that related Chinese websites may take 0.5\% of the whole Chinese pre-training dataset of GPT-4o
\footnote{Project website: \url{https://pollutedtokens.github.io}}.

\section{Preliminaries}

\noindent \textbf{Tokenization}. This stands as a cornerstone in natural language processing (NLP), where raw textual data is segmented into fundamental units called tokens \cite{choo2023study,vijayarani2016text,grefenstette1999tokenization}. For instance, for a continuous text sequence ``Words can be one token or not: indivisible'', advanced GPT's tokenizer yields \{``Words'', `` can'', `` be'', `` one'', `` token'', `` or'', `` not'',  ``:'', `` indiv'', ``isible''\}.

Among various tokenization methods such as WordPiece \cite{song2020fast,wu2016google}, SentencePiece \cite{hellsten2024incremental,kudo2018sentencepiece} and ULM \cite{wang2021multi,kudo2018subword}, Byte-Pair Encoding (BPE) \cite{wang2020neural,sennrich2015neural} emerges prominently. 
It first splits training text into words, a process called pretokenization (e.g., splitting on whitespace). Then, words are split into bytes to form the starting vocabulary. BPE iteratively counts the frequency of each neighboring pair of tokens and picks the most frequent one to merge, adding the merge rule and the merged token to the merge list and the vocabulary. This continues until the desired vocabulary size is reached. To tokenize a text sequence, BPE tokenizer splits the text into bytes and applies the learned merge rules. Therefore, the vocabulary of the tokenizer reflects rich distributional information about the training corpus \cite{hayase2024data,xu2024bridging,weber2023evaluating}.

\noindent \textbf{Chinese language and characters.} Chinese language is a complex system that relies on individual characters as the basic building blocks \cite{defrancis1986chinese,wang1973chinese,morrison1815grammar}. 
Unlike phonetic alphabets, each Chinese character usually does not convey a specific meaning but serves as a symbolic representation that carries semantic potential \cite{williams2010chinese,dai2007chinese,liu2004online}. They only convey full meaning when appearing with more characters. 
This makes Chinese language context-dependent \cite{yang2013word,hsieh2012improving,wu2007context}: \textit{the meaning of a single Chinese character often shifts or becomes more specific through its association with other Chinese characters in multi-character compounds}. 
For example, for the \poc token ``毛片'' (``pornographic film''), each of its character ``毛'' (``wool''), ``片'' (``film'') is normal.
Due to this context-dependency, it is challenging to detect \poc tokens from LLMs' vocabularies.

\noindent \textbf{Abnormal tokens.} Recent research has identified various types of abnormal tokens within LLMs' vocabularies, including glitch tokens and under-trained tokens. Glitch tokens are abnormal tokens that can trigger unpredictable or nonsensical outputs, diverging from human normative responses \cite{geiping2024coercing,fell2023search}. \citet{li2024glitch} conduct a comprehensive and systematic empirical study on the glitch token phenomenon in LLMs, including taxonomy and detection methods. Under-trained tokens are those present in the tokenizer vocabulary but nearly or fully absent during model training, leading to unwanted model behavior \cite{land2024fishing}. \citep{watkins2023solidgoldmagikarp,rumbelow2023solidgoldmagikarp} have identified these tokens through model and tokenizer analysis. \citet{land2024fishing} provide an automated tool for detection based on the model embedding weights and tokenizer configuration.

\citet{polluGpt4o} finds that GPT-4o's vocabulary is polluted by Chinese Internet scams, such as pornography or online gambling websites. This paper builds upon all these studies and focuses on \poc tokens.

\section{Polluted Chinese (\poc) tokens in GPT}

We first formalize the definition and taxonomy of \poc tokens, then demonstrate that GPT cannot understand them, although they are GPT's tokens.

\subsection{Definition and taxonomy}
\label{sec:defTax}

\poc tokens are sourced from illegal websites in Chinese involving porn or online gambling. However, it is difficult to give a definition and taxonomy for these tokens due to their incompatibility with mainstream Chinese linguistics. 

To overcome this challenge, we assemble an interdisciplinary research team with 6 experts owning PhD degrees in philosophy, sociology, Chinese linguistics, and computer science. 
In collaboration with this expert panel, our formal definition of the \\
\noindent\textbf{polluted Chinese tokens (\cpt)} is: 
\ul{\textit{Chinese tokens from LLM's vocabularies that encode undesirable, uncommon, or useless content (i.e., 3U principle) from the perspective of current mainstream Chinese linguistics.}}

Among the 3U principle, undesirable content is inappropriate, unethical, or violates legal regulations, such as pornography and online gambling content, e.g., ``波*野结衣'' (``Yui Hatano''); uncommon content is unlikely to appear within standard Chinese linguistic contexts, e.g., ``大香蕉'' (``big banana''); useless content lacks meaningful linguistic or semantic value in Chinese corpus processing, e.g., ``给主人留下些什么吧'' (``leave something for the master''). The presence of these tokens indicates a significant pollution in the Chinese language portion of training data.
Then, the expert panel further establishes a taxonomy:
\begin{packeditemize}
\item \textbf{Adult content} contains explicit or implicit sexual references, such as ``波*野结衣'' (``Yui Hatano'')\footnote{In \autoref{app:expertLabeling}, we discuss why ``波*野结衣'' is an adult content token rather than a name token.}, ``青青草'' (``green grass'').
\item \textbf{Online gambling} refers to gambling websites, betting platforms, lotteries, or related gambling activities, such as ``天天中彩票'' (``everyday lottery''), ``菲律宾申博''  (``Philippine sunbet'').
\item \textbf{Online game} is related to unofficial or unauthorized online game services, such as ``传奇私服'' (``legend private server'').
\item \textbf{Online video} is related to online video platforms or streaming content, such as ``在线观看'' (``watch online''), ``免费视频'' (``free video'').
\item \textbf{Anomalous} represents rare, peculiar, or contextually irrelevant phrases, such as ``给主人留下些什么吧'' (``leave something for the master'').

\end{packeditemize}
Based on this taxonomy, our team labels all Chinese tokens from advanced GPT's vocabulary (see details of our labeling pipeline and members' backgrounds in \autoref{app:expertLabeling}), which could serve as labels for fine-tuning an LLM for \cpt detection in \autoref{sec:PolluDetect}. The labeling results are shown in the first row of \autoref{tab:CPTRatio}.

\subsection{\cpt cause GPT's weird outputs}

It is indicated that \cpt can cause weird outputs for GPT-4o when released\footnote{\url{https://github.com/openai/tiktoken/issues/297}}. 
However, it is disappointing that today's 4o and more advanced 4.5, 4.1 models still suffer the same issue.  
To further investigate and verify this issue, we perform two tasks to evaluate how GPT comprehends these \poc tokens in comparison to normal ones. 

\begin{figure*}[t]
    \centering
    \includegraphics[width=0.9\linewidth]{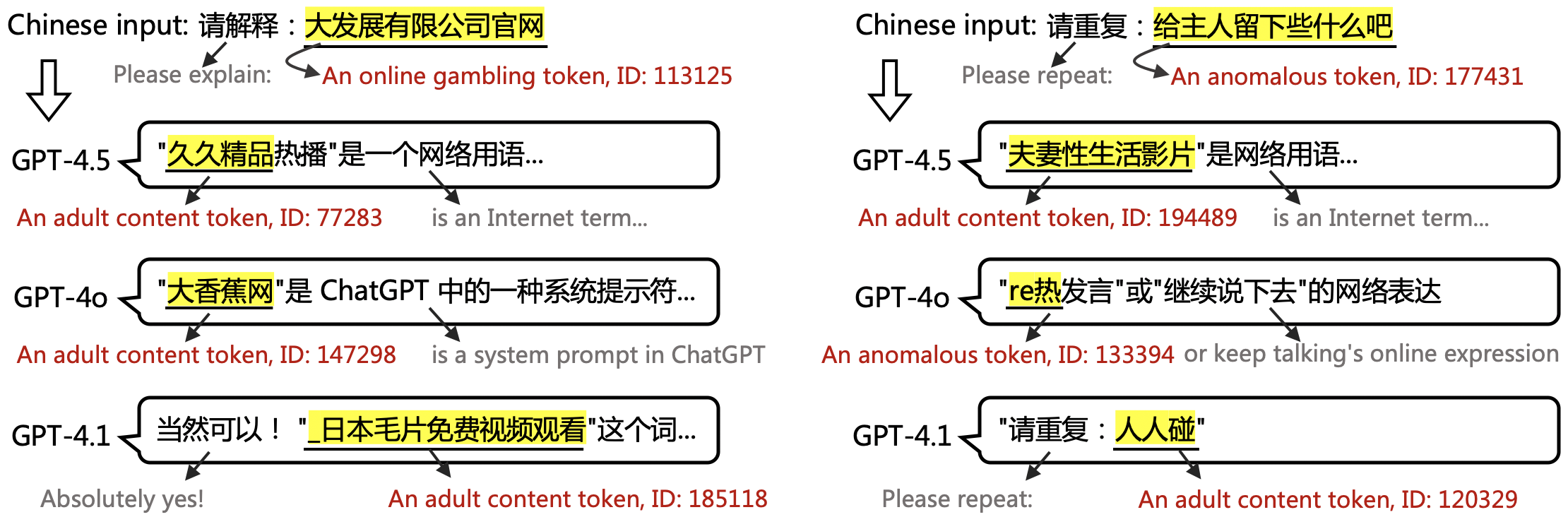}
    \vspace{-1ex}
    \caption{An example: the outputs of GPT-4.5, 4.1, and 4o when the input involves \cpt. We use Chinese to query and list the \translate{corresponding translation in gray color to help reading}.  GPT cannot explain or repeat \cpt. Full outputs are in \autoref{app:degradeExpLabeling}, and please kindly note GPT has different but similar outputs if reproduced.}
    \label{fig:examplesDegradeInference}
    \vspace{-2ex}
\end{figure*}
 
\begin{packeditemize}
    \item \textbf{Interpretation task} aims to measure GPT's comprehension of \poc tokens \cite{edman2024cute}. We use a prompt template ``Please explain: \{Token\}'' to assess whether the GPT knows the semantic meaning of the token. 
    \item \textbf{Repetition task} aims to measure GPT's external generation capability of \cpt \cite{xue2023repeat}. We use a prompt template ``Please repeat: \{Token\}'' to examine whether the LLM can reproduce the exact tokens.
\end{packeditemize}

These two tasks assess the internal comprehension and external generation capability of LLMs concerning \cpt. The former is measured by DeepSeek-V3 \cite{liu2024deepseek} with strong Chinese capabilities and contains no advanced GPT's \cpt in its vocabulary; the latter is measured by verbatim match \cite{ippolito2022preventing}. 

\begin{table}[t]
\centering
\setlength{\tabcolsep}{8pt}
\renewcommand{\arraystretch}{0.75}
\resizebox{\linewidth}{!}{
\begin{tabular}{@{}c|cccc@{}}
\toprule
 &  \multicolumn{2}{c}{Interpretation} & \multicolumn{2}{c}{Repetition}  \\
 & Normal & \poc & Normal & \poc \\
\midrule
4-turbo & 86.1 & 72.2\diffdown{13.9} & 100.0 & 100.0 \\
\midrule
4o & 88.0 & 43.7\diffdown{44.3} & 96.0 & 54.2\diffdown{41.8} \\
4o-mini & 84.8 & 31.2\diffdown{53.6} & 91.0 & 35.4\diffdown{55.6} \\
o1-mini & 85.6 & 28.9\diffdown{56.7} & 94.3 & 46.7\diffdown{47.6} \\
o3-mini & 89.0 & 30.7\diffdown{58.3} & 89.6 & 33.1\diffdown{56.5} \\
o4-mini & 91.4 & 39.7\diffdown{51.7} & 92.6 & 36.8\diffdown{55.8} \\
4.5-preview & 81.8 & 47.9\diffdown{33.9} & 97.1 & 60.3\diffdown{36.8} \\
4.1 & 90.2 & 46.2\diffdown{44.0} & 96.7 & 51.7\diffdown{45.0} \\
4.1-mini & 88.6 & 34.5\diffdown{54.1} & 94.0 & 39.1\diffdown{54.9} \\
4.1-nano & 86.3 & 33.7\diffdown{52.6} & 92.9 & 46.0\diffdown{46.9} \\
\bottomrule
\end{tabular}
}\vspace{-1ex}
\caption{Accuracy (\%) of interpretation and repetition tasks. Advanced GPT models hardly understand \cpt compared to normal ones. Instead, GPT-4-turbo works well as its vocabulary is not polluted (\autoref{tab:CPTRatio}).
}\vspace{-3ex}
\label{tab:degradeInference}
\end{table}

\autoref{fig:examplesDegradeInference} shows an example for each task on GPT-4o. It cannot explain the meaning of \cpt but instead generate nonsense. Moreover, it cannot repeat \cpt but generate irrelevant content. Such a deficiency is inherited by advanced GPT models because they share the same vocabulary. \autoref{tab:degradeInference} statistically shows that \cpt cause approximately 50\% decrease compared to normal Chinese tokens. Instead, GPT-4-turbo, whose vocabulary is not polluted (shown in \autoref{tab:CPTRatio}), can almost explain and perfectly repeat \cpt.  

\subsection{Input \poc tokens, output \poc tokens, why?} 
\label{sec:whyIOpoc}

Our assumption is that \poc related contents widely exist in the pre-training dataset but are not under-trained during the later training stage \cite{land2024fishing,li2024glitch}, thereby degrading the GPT's generation on these \poc tokens.  

We briefly verify our assumption with the mC4 corpus~\cite{xue2020mt5}. 
We pick one news website from mC4 and then use the GPT's vocabulary as the tokenizer to analyze. 
It is surprising that even for this news website in mC4, GPT's \poc tokens appear repeatedly and are concentrated at this web's very beginning and end. 
This may verify our assumption that the \poc tokens consistently appear in sequence in the pre-training datasets, which creates associations among them during the pre-training phase. 
But \poc tokens are not explicitly trained in subsequent phases. When \poc tokens are input, the model tends to output other related \poc tokens.

\begin{figure}[t]
    \centering
    \includegraphics[width=0.92\linewidth]{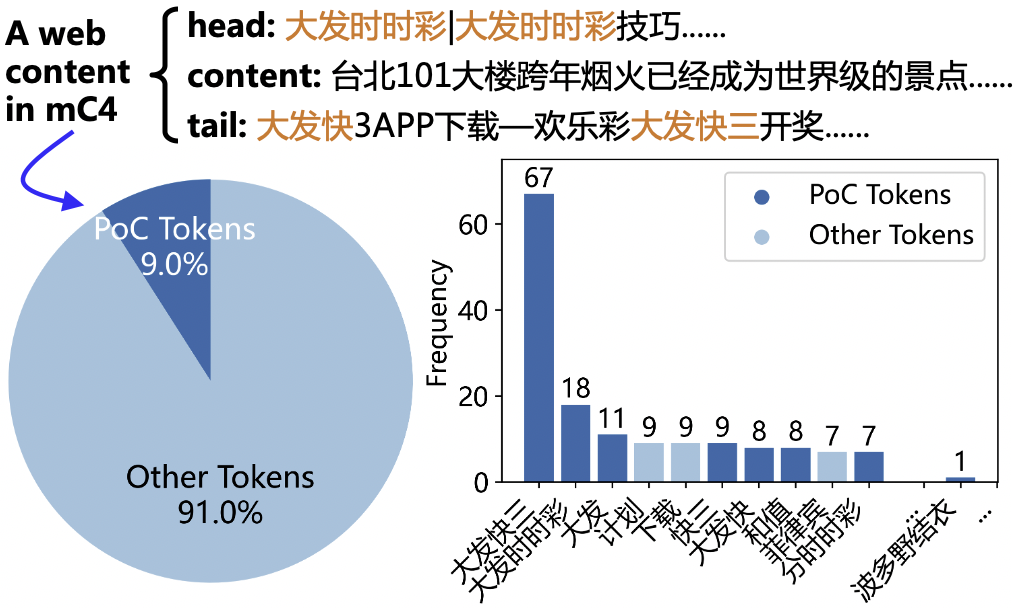}
    \vspace{-1ex}
    \caption{A Chinese news website in mC4: most of its contents are normal, but polluted contents are at the head and tail of the web (GPT's \poc token ``大发快三'' appears 67 times). More details are in \autoref{app:mc4}.}
    \label{fig:pollution}
    \vspace{-3ex}
\end{figure}

\begin{tcolorbox}[colback=blue!5!white,colframe=gray!75!black,left=1mm, right=1mm, top=0.5mm, bottom=0.5mm, arc=1mm]
    \textbf{Observation 1}: GPT's \poc tokens encode information of adult content, gambling, etc., but GPT cannot understand them. The reason may be that these \poc tokens are concentrated in corpus only for pre-training.
\end{tcolorbox}

\section{Detect \poc tokens in more vocabularies}

\begin{table*}[t]
\centering
\setlength{\tabcolsep}{8pt}
\renewcommand{\arraystretch}{0.75}
\resizebox{\linewidth}{!}{
\begin{tabular}{@{}c|ccccc|c@{}}
\toprule
 & Adult Content & Online Gambling & Online Game  & Online Video & Anomalous & Total\\
\midrule
GPT-4o/o1/o3/4.5/4.1/o4 & 219~(13.2\%) & 459~(27.7\%) & 14~(0.84\%) & 47~(2.83\%) & 34~(2.05\%) & 773~(46.6\%) \\
BLOOM & 8~(0.11\%) & 0~(0.00\%) & 4~(0.06\%) & 0~(0.00\%) & 106~(1.51\%) & 118~(1.68\%) \\
Qwen2/2.5/3 & 1~(0.02\%) & 13~(0.27\%) & 26~(0.54\%) & 1~(0.02\%) & 7~(0.15\%) & 48~(1.00\%) \\
GLM4 & 4~(0.05\%) & 2~(0.03\%) & 6~(0.08\%) & 2~(0.03\%) & 5~(0.07\%) & 19~(0.25\%) \\
DeepSeek-V3/R1 & 6~(0.06\%) & 0~(0.00\%) & 2~(0.02\%) & 1~(0.01\%) & 8~(0.08\%) & 17~(0.17\%) \\
MiniCPM & 0~(0.00\%) & 2~(1.92\%) & 0~(0.00\%) & 2~(1.92\%) & 2~(1.92\%) & 6~(5.77\%) \\
LLaMA-3/3.1/3.2 & 0~(0.00\%) & 2~(1.92\%) & 0~(0.00\%) & 2~(1.92\%) & 2~(1.92\%) & 6~(5.77\%) \\
Gemma-1/2 & 0~(0.00\%) & 0~(0.00\%) & 1~(0.08\%) & 0~(0.00\%) & 0~(0.00\%) & 1~(0.08\%) \\
GPT-4/4-turbo/3.5 & 0~(0.00\%) & 0~(0.00\%) & 0~(0.00\%) & 0~(0.00\%) & 0~(0.00\%) & 0~(0.00\%) \\
\bottomrule 
\end{tabular}
}
\vspace{-1ex}
\caption{Number (Ratio $\%$) of \poc tokens in LLMs' Chinese vocabularies (one token containing more than 2 Chinese characters).
}
\label{tab:CPTRatio}\vspace{-1ex}
\end{table*}

We also aim to explore the \poc tokens in other LLMs. 
Please note that GPT cannot understand these \poc tokens, so prompting GPT to read other LLMs' vocabularies does not work. 
Therefore, we design \detector to automatically label tokens. It is fine-tuned from a Chinese LLM which has many Chinese tokens in the vocabulary. 

\subsection{\detector: LLM for detection}
\label{sec:PolluDetect}

Leveraging the expert labeling results on advanced GPT's vocabularies, we fine-tune a GLM-4-32B \cite{glm2024chatglm}, due to its good comprehension ability of Chinese and less polluted vocabulary (shown in \autoref{tab:CPTRatio}), to develop \detector for \poc token detection and classification.

\cpt can be subtle or implicit, not directly showing their nature in terms of semantics. Therefore, detecting \cpt often requires contextual information. 
Considering such a characteristic, we implement a web-browsing mechanism in \detector, following \cite{vu2023freshllms}. Specifically, we utilize the SerpApi\footnote{\url{https://serpapi.com/}} to retrieve the top 10 Google search results for the token to evaluate, especially their snippet information. We then incorporate them into the prompt as contextual information. The detection prompt template is as follows.
\begin{tcolorbox}
    [width=\linewidth,colback={white},title={\fontsize{9.5}{7}\selectfont Prompt template of \detector},coltitle=white,left=1pt,right=1pt,top=1pt,bottom=1pt] 
{\small
I am analyzing the Chinese token \placeholder{Token} from LLMs' vocabulary. Please categorize it based on the provided taxonomy and the Google search results for this token.\\
The taxonomy is as follows：\\
\placeholder{Taxonomy}\\
The search engine results are as follows：\\
\placeholder{Search engine result}\\
The pipeline of analysis is as follows：\\
\placeholder{Pipeline of analysis}\\
Please categorize the Chinese token：\placeholder{Token}\\
Please only output the category：
}
\end{tcolorbox}
The fine-tuning labels are expert annotations in \autoref{sec:defTax}. Since we focus on detecting \poc tokens, we use Chinese prompts (\autoref{app:detectorDetail}).

\subsection{\cpt within LLMs vocabularies}

\autoref{tab:CPTRatio} shows the \cpt detected by \detector in 9 vocabularies of 23 LLMs, except for GPT, which are labeled by our expert panel. \cpt widely exist in various LLMs' vocabularies.
Conversely, the vocabularies of GPT-4/4-turbo/3.5 contain no \cpt, which may indicate a clean training corpus. 
Among the detected \cpt, adult content, online gambling, and anomalous content are the majority. 
This yields the significance of data cleaning on these contents. We show the detected \cpt in \autoref{app:LLMsCPT}.

\begin{tcolorbox}[colback=blue!5!white,colframe=gray!75!black,left=1mm, right=1mm, top=0.5mm, bottom=0.5mm, arc=1mm]
    \textbf{Observation 2}: \poc tokens widely exist in contemporary LLMs' vocabularies, especially in GPT, BLOOM, and Qwen.
\end{tcolorbox}

\section{Estimate training data pollution} 

Since the widely used BPE tokenizer \cite{sennrich2015neural} is originated from the field of data compression \cite{gage1994new}, tokens generated through BPE naturally reflect the statistical distribution of the training corpus. Leveraging BPE vocabularies, \citet{hayase2024data} estimate the mixture ratios of different data sources, but not the specific frequency of certain tokens in training corpus. The main challenge is that the training corpus is too large and complex, causing difficulty in estimating a certain token (as mentioned in \autoref{sec:intro}). Therefore, we design \trace to estimate the frequency of \cpt, revealing the severity of Chinese training data pollution.

\subsection{\trace: trace Chinese data pollution}

\trace provides fine-grained investigation to reveal the presence frequency of specific \cpt in training corpus. By examining tokens individually, we pinpoint which one contributes most significantly to data pollution. The aggregated results of all polluted Chinese tokens can reveal the holistic scale of Chinese training data pollution.

\noindent \textbf{From token ID to frequency.} The main idea of estimation is simple yet effective. Inspired by Zipf's law \cite{piantadosi2014zipf,saichev2009theory}, which states that the frequency of a word in a natural language corpus is approximately inversely proportional to its frequency rank, we aim to fit the relationship between token IDs from the tokenizer and tokens' frequencies. With this fitted relationship, we can estimate a token's proportion in the training corpus directly from its token ID. 

\begin{figure}
    \centering
    \includegraphics[width=\linewidth]{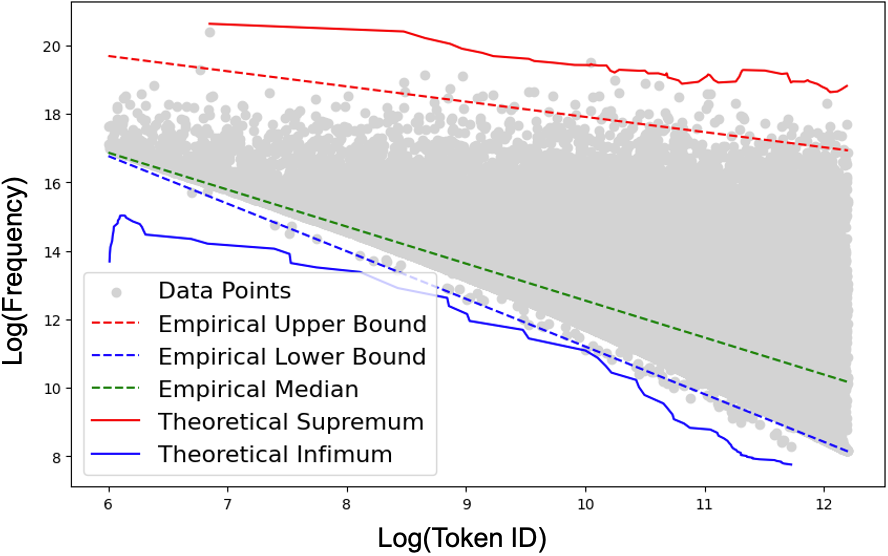}
    \caption{Data points are confirmed to fall within the theoretical supremum and infimum, and thus can be reliably estimated using empirical bounds and median.}
    \label{fig:bound}
\end{figure}

Specifically, we first train a BPE tokenizer on an open-source corpus (e.g., Pile \cite{gao2020pile}, C4 \cite{raffel2020exploring}) and count the frequency of each token from the resulting vocabulary. After obtaining all frequency–ID pairs, we apply a logarithmic transformation to both frequency and token ID, converting the inverse proportionality described by Zipf's law into a linear relationship. We then plot all data points in a scatter plot to perceive the data distribution, as illustrated in \autoref{fig:bound}. 

We observe that the data points do not align along a perfect linear distribution. This is due to the inherent complexity and diversity of natural language training corpora. However, their upper and lower boundaries approximately exhibit linear relationships. Consequently, we can empirically derive the upper and lower bounds for this fitted relationship, which allow us to estimate the frequency range of tokens in the training corpus solely based on their token IDs. If a precise estimation is required rather than a range, we can also derive an empirical median from the data distribution. Additionally, we theoretically derive the supremum and infimum of this fitted relationship from the BPE algorithm itself, thus validating the reasonableness of our empirical estimation.

\noindent \textbf{Empirical estimates of upper bound, median, and lower bound.} Inspired by quantile regression \cite{romano2019conformalized,koenker2005quantile}, which captures the distributional trends of extreme values, we fit the empirical upper and lower bounds by applying asymmetric penalty weights to data points. The loss function of quantile regression is as follows:
\begin{equation}
\min_{\beta} \sum_{(x, y)} \rho_{\tau}(y - x^{\top}\beta),
\end{equation}
where $\beta$ is the regression coefficient to estimate, $(x,y)$ is a data point, and $\rho_{\tau}(\cdot)$ is:

\begin{equation}
\rho_{\tau}(u) = 
\begin{cases}
\tau u, & u \geq 0 \\
(\tau - 1) u, & u < 0
\end{cases},
\end{equation}
where $\tau$ is the quantile parameter, which applies asymmetric penalty weights $ (0<\tau<1)$ to ensure approximately $\tau$ (resp., $1-\tau$) of data lie below (resp., above) the fitted regression line. By properly selecting $\tau_{min}$ and $\tau_{max}$ (e.g., 0.01 and 0.999), we adjust the empirical upper and lower bounds of the fitted frequency-token ID relationship. $\tau_{med}=0.5$ naturally determines the empirical median.

Once we get the regression coefficient $\beta_{min}$, $\beta_{med}$, $\beta_{max}$, based on $\tau_{min}, \tau_{med}, \tau_{max}$, the empirical lower bound, median, and upper bound can be represented as (also plotted in \autoref{fig:bound}):
\begin{equation}
F_{*}(t_i) = e^{\beta_{*}^{(1)}} \times i^{\beta_{*}^{(2)}}, * \in \{\text{min}, \text{med}, \text{max}\},
\end{equation}
where $t_i$ denotes the token of tokenID $i$ and $F(\cdot)$ denotes its frequency.

\noindent \textbf{Theoretical supremum and infimum.} The nature of the BPE algorithm inherently assigns smaller (resp., larger) token IDs to tokens with higher (resp., lower) frequencies. Moreover, since each new token is constructed from existing tokens in the vocabulary, we argue that the frequency of any token cannot exceed the minimum frequency among all of its constituent subtokens, nor can it be lower than the maximum frequency among all tokens that contain it as a subtoken. Such a relationship is formally described by the following equation:
\begin{equation}
\begin{aligned}
&\max_{\substack{t_i \subseteq_{\text{sub}} t_j}} F(t_j) \leq F(t_i) \leq \min_{\substack{t_k \subseteq_{\text{sub}} t_i}} F(t_k),
\end{aligned}
\end{equation}
where $\subseteq_{\text{sub}}$ denotes substring.

The argument above establishes theoretical upper and lower bounds. To confirm that these bounds are the supremum and infimum, we need to demonstrate that they are the smallest upper bound and the largest lower bound. We employ a proof by contradiction: we construct a naive training corpus ``ab ab''. For the token ``ab'', its frequency $F(``ab")=2$ and its upper bound $\min_{\substack{t_k \subseteq_{\text{sub}} ``ab"}} F(t_k)=\min\{F(``a"),F(``b")\}=\min\{2,2\}=2$. Suppose, for contradiction, that $\min_{\substack{t_k \subseteq_{\text{sub}} ``ab"}} F(t_k)$ is not the smallest upper bound, this would imply the existence of another upper bound strictly smaller than $\min_{\substack{t_k \subseteq_{\text{sub}} ``ab"}} F(t_k)=2$ yet still greater than or equal to $F(``ab")=2$, leading to a contradiction; Similarly, for the token ``a'', the frequency $F(``a")=2$ and its lower bound $\max_{\substack{``a" \subseteq_{\text{sub}} t_j}} F(t_j)=\max\{F(``ab")\}=2$. Assuming $\max_{\substack{``a" \subseteq_{\text{sub}} t_j}} F(t_j)$ is not the largest lower bound would imply the existence of another lower bound strictly greater than $\max_{\substack{``a" \subseteq_{\text{sub}} t_j}} F(t_j)=2$ yet still smaller than or equal to $F(``a")=2$, again resulting in contradiction. Hence, these bounds indeed represent the supremum and infimum. 

We also plot the theoretical supremum and infimum in \autoref{fig:bound}, which indicates that data points are confirmed to fall within supremum and infimum, and thus can be reliably estimated using empirical bounds and median.

\subsection{Estimation results}
\label{sec:resEstimation}

We first estimate Chinese data pollution on an open-sourced dataset mC4 \cite{xue2020mt5} with English and Chinese corpus to verify the accuracy of \trace. We use the average empirical median and upper/lower bounds derived from other 4 open-sourced datasets to estimate mC4.

\smallskip
\noindent \textbf{Similarity of \trace between different training corpora.} The above approach only works if the empirical estimates derived from one corpus can be transferred to another corpus. To verify this, we prepare a Chinese pretraining corpus by mixing the related webpages from CommonCrawl\footnote{\url{https://commoncrawl.org/}} of 200 normal Chinese tokens and 10 \cpt for each \poc token category. Then we mix the Chinese pretraining corpus with 4 open-source pretraining corpora with a mix ratio of 10\% following. Such construction of corpus is due to the rarety of accessible Chinese pretraining corpus other than mC4. For this constructed corpus, we can compute the frequencies of any tokens, which serve as the ground truth to verify empirical estimates. We then train a BPE tokenizer on this mixed corpus to get the vocabulary, and estimate the frequency of each \poc token in the vocabulary by empirical estimates derived from another mixed corpus.

\begin{table}[t]

\label{tab:transferCPoTrace}
\centering
\setlength{\tabcolsep}{8pt}
\renewcommand{\arraystretch}{0.75}
\resizebox{0.8\linewidth}{!}{
\begin{tabular}{@{}c|cccc@{}}       
\toprule
& \multicolumn{4}{c}{Estimate from} \\
To &  Pile & C4 & Dolma & Roots \\
\midrule
Pile & 99.6 & 99.8 & 99.6 & 96.1 \\
C4 & 99.7 & 99.8 & 99.7 & 99.7 \\
Dolma & 59.9 & 68.6 & 99.7 & 61.9 \\
Roots & 99.8 & 99.8 & 99.8 & 99.7 \\
\bottomrule 
\end{tabular}
}
\vspace{-1ex}
\caption{Accuracy (\%) of whether the frequency lie in the empirical bounds estimated by \trace. \trace effectively estimate \cpt frequency in pretraining corpus, and is transferable to other corpus. 
}
\label{tab:transferCPoTrace}
\end{table}

\autoref{tab:transferCPoTrace} shows that our estimation transfers well between 4 open-source training corpora: Pile \cite{gao2020pile}, C4 \cite{raffel2020exploring}, Dolma \cite{dolma}, and Roots \cite{laurenccon2022bigscience}. The accuracy of whether the token's frequency of one corpus falls within empirical bounds estimated from another corpus is high in almost all cases. When the corpus to estimate is the same as the one to derive empirical bounds, it means we estimate from the original corpus to the corpus mixed by the constructed Chinese pretraining corpus. Among the four tested corpora, Dolma is slightly harder to estimate because its data distribution is slightly different (shown in \autoref{app:distributionOpenCorpus}).

\begin{figure}[t]
    \centering
    \includegraphics[width=0.85\linewidth]{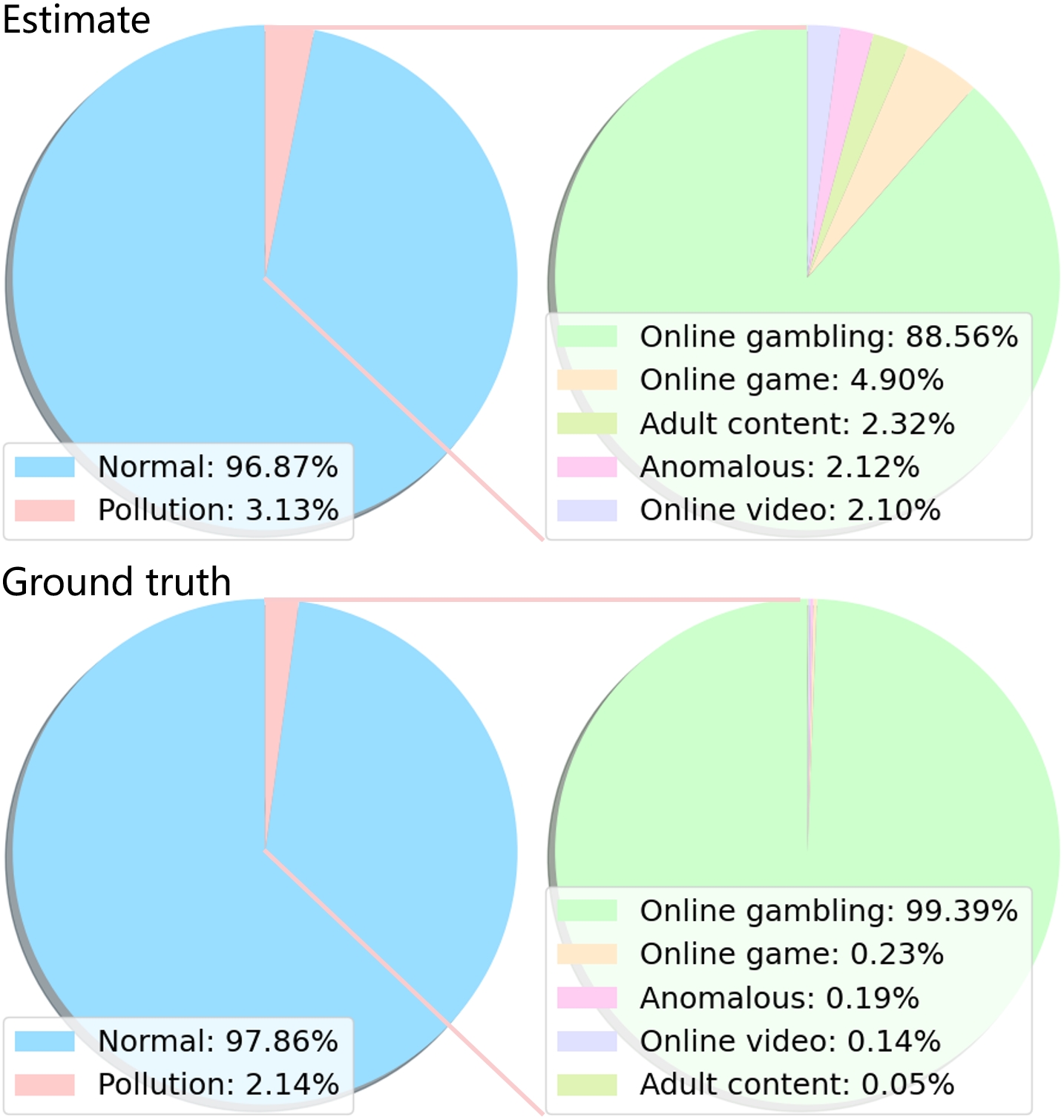}
    \caption{Estimated polluted ratio of Chinese training corpus within mC4 compared to ground truth (with each token manually verified in mC4).}
    \label{fig:estimatedRatiomC4}
    \vspace{-1em}
\end{figure}

\smallskip
\noindent \textbf{Estimation of mC4.} Since we demonstrate the transferability of \trace between different training corpora, we can estimate mC4 as follows: leveraging the vocabulary of the tokenizer trained on a random subset of mC4, we use \detector to identify \cpt; then, we use the average empirical median (i.e., $F_{\text{med}}$) derived from the 4 open-sourced corpora to estimate each token's frequency; consequently, the ratio for each content category (i.e., $R(\mathcal{C})$) can be expressed as: 

\begin{equation}
R(\mathcal{C})=\frac{\Sigma_{t_i\in \mathcal{C}}F_{\text{med}}(t_i)S(t_i)}{\Sigma_{t_i\in \text{CN}}F_{\text{med}}(t_i)S(t_i)},
\label{eq:ratio}
\end{equation}
where $S(\cdot)$ is the size of the token (3 bytes for 1 Chinese character), CN represents Chinese tokens.

\autoref{fig:estimatedRatiomC4} shows the estimated results compared to the ground truth on mC4. We observe that 3.13\% of Chinese data is polluted, which is comparable to the ground truth value 2.14\%. However, estimating the distribution within pollution is more difficult because it is highly imbalanced, which can lead to possible outliers. In short, \trace is acceptable to estimate overall pollution.

\subsection{Speculate GPT-4o's ``波*野结衣'' content}

``波*野结衣'', appearing as a token in GPT's vocabulary, is the Chinese name of a famous Japanese pornstar Yui Hatano. 
This is one of the few names in Chinese that become GPT's tokens while others are ``特朗普'' (Donald Trump's Chinese name, ID: 161,031), ``五月天'' (a famous Chinese rock band, ID: 45685), etc. 
We have no clear idea why she is the only pornstar whose Chinese name becomes a token of GPT. 
However, we pick this token as an example because its subsequences (``*野结衣'', ``*野结'', ``*野'') are also GPT's tokens. 
We determine that these four tokens are only related to ``波*野结衣'' which gives us an opportunity to speculate GPT-4o's ``波*野结衣'' content.

The key insight is as follows. 
\textit{First, we estimate a ratio of ``波*野结衣'' in the training dataset. 
Then, we use ``波*野结衣'' related websites to mix with an open-sourced dataset with this estimated ratio and generate a vocabulary via BPE. 
If this ratio is correct, all four tokens' appearance (ID) should be very similar to GPT's.}

We first leverage \trace to speculate. 
Using GPT-4o's token ID 185,946 for ``波*野结衣'', \autoref{eq:ratio} yields an estimated token ratio of 0.000085\%. 
To get the related web content ratio, we find the polluted webpages containing ``波*野结衣'' within CommonCrawl and compute its presence ratio $R_p$. The related web content ratio is therefore $0.000085\%/R_p=0.5\%$. 

\begin{figure}[t]
    \centering
    \includegraphics[width=\linewidth]{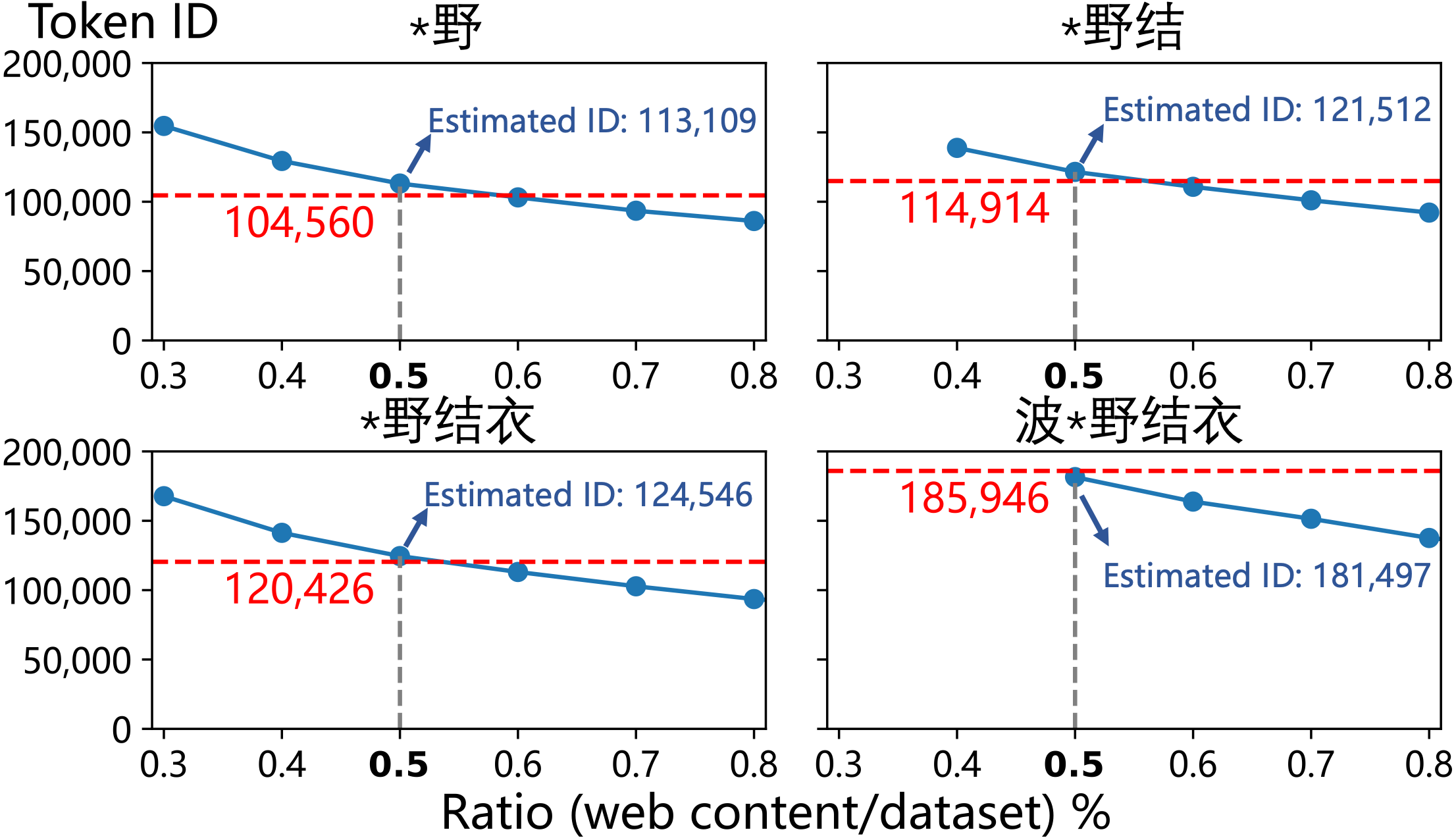}  
    \vspace{-2em}
    \caption{Mixing ``波*野结衣'' related webpages with Pile at our estimated ratio (0.5\%) can reproduce \textcolor{red}{GPT-4o's token ID} of ``波*野结衣'' and its subsequences.}
    \label{fig:reproduce}
    \vspace{-1em}
\end{figure}

To verify the above estimation, we mix the webpages related to ``波*野结衣'' from CommonCrawl to Pile, and perform BPE tokenization to observe the token IDs of the four tokens. 
\autoref{fig:reproduce} shows that the reproduced token ID is close to that of GPT-4o's vocabulary (181,497 compared to 185,946), as well as for the subsequences (``*野结衣'', ``*野结'', ``*野'') simultaneously, while a more or less ratio leads to clear different results. 

Moreover, since the token ID directly corresponds to token frequency within train corpus via \autoref{eq:ratio}, we surprisingly observe that ``波*野结衣'' appears $\sim2.6\times$ more often than ``您好'' (ID: 188,633, translating as ``Hello''), despite the latter being undoubtedly more common in daily usage. This indicates there may be a gap between GPT-4o's training dataset and Chinese language, which may degrade its Chinese capability \cite{lin2025prompting,glm2024chatglm,wen2024chinese}. 
We give more GPT's \cpt in \autoref{app:GPT4oCPT}.

\begin{tcolorbox}[colback=blue!5!white,colframe=gray!75!black,left=1mm, right=1mm, top=0.5mm, bottom=0.5mm, arc=1mm]
    \textbf{Observation 3}: GPT-4o's training dataset may contain a significant ratio of polluted Chinese contents, e.g., ``波*野结衣'' ("Yui Hatano") appears $\sim2.6\times$ than ``您好'' ("hello"), and its related webpages takes $\sim0.5\%$. 
\end{tcolorbox}

Please kindly note that we are not from OpenAI, so we have no way to verify this speculation. 
However, since OpenAI's pre-training data for GPT-4o is also originated from the Internet so it is likely that it shares a similar distribution with open-sourced ones like Pile. 
In the end, we hope this speculation can be verified in the future and these Chinese polluted contents can be effectively reduced in LLMs' training datasets.  

\section{Conclusion}

In this paper, based on the GPT's \poc tokens, we first perform a rigorous labeling on \poc tokens in GPT's vocabulary. 
Then, we build a detector to locate \poc tokens in 9 vocabularies of 23 LLMs. 
We also study how to speculate the Chinese training data pollution via the \poc token's appearance (ID).
\poc tokens exist widely, reflecting the serious Chinese data pollution in LLM training. 

\clearpage
\section*{Limitations}

\noindent \textbf{Close-source of GPT-4o training data.} Since our proposed \trace can estimate Chinese data pollution via LLMs' vocabulary, it is feasible to estimate Chinese data pollution via GPT-4o's vocabulary. However, we have no ground truth to verify this estimation due to the close-sourced GPT-4o training data. We hope to verify the estimation when the train corpus is accessible one day.

\smallskip
\noindent \textbf{Polluted tokens in other languages.} Since we focus on Chinese data pollution, our work contains no investigation of polluted tokens in other languages. Extending the research scope to other languages requires a larger expert panel with multilingual capability, which is currently a challenge for us. 
However, data pollution and polluted tokens for other languages did exist. 
For instance, it is reported that Korean tokens have the similar issue\footnote{\url{https://www.technologyreview.com/2024/05/17/1092649/gpt-4o-chinese-token-polluted/}}. 
We hope more language experts can pay attention to this pollution issue and give more investigation.

\smallskip
\noindent \textbf{Readability to non-native Chinese readers.} 
In this paper, there are extensive Chinese characters used and we did our best to translate most of them for general readability. 
However, it is worth noted that many of those tokens are hard to translate even for linguistic experts since they are not reading via semantics. 
We would like to emphasize that this paper does not study the linguistic and expression but just aims to draw attention to the training data pollution in many SOTA LLMs by using Chinese as an example.
In summary, we did try our best to translate necessary part into English for better readability in general.
We sincerely hope more researchers can join in this research direction that will improve future works' readability. 

\section*{Ethics Statement}
ACL Ethics Policy is respected in this work. This work studies polluted Chinese tokens within LLMs vocabulary and polluted content within Chinese train corpus. We investigate open-sourced LLMs vocabulary and train corpus whose terms, conditions, and copyright are respected. This paper may include offensive and upsetting content which need to be use with caution for future research.

We adhere strictly to the Association for Computational Linguistics (ACL) guidelines on responsible NLP research\footnote{\url{https://aclrollingreview.org/responsibleNLPresearch/}}, ensuring compliance with copyright and ethical standards.
For instance, the portrait of Yui Hatano (“波*野结衣”)\footnote{\url{https://en.wikipedia.org/wiki/File:Yui_Hatano,2016(cropped).jpg}}, referenced in~\autoref{fig:first}, is publicly available under a Creative Commons Attribution 4.0 International (CC BY 4.0) license.
This license permits the use of the image for research and academic purposes.

\section*{Acknowledgements}
This work was supported by the Key Program of the National Natural Science Foundation of China under No. 62132011, the National Science Foundation for Distinguished Young Scholars under No. 62425201, Ant Group, and the Center of High Performance Computing, Tsinghua University.

\bibliography{main}
\bibliographystyle{acl_natbib}

\clearpage
\appendix

\section{Labeling tokens in GPTs' vocabulary}
\label{app:expertLabeling}

\subsection{Labeling process and system}

\begin{figure}[!htbp]
    \centering
    \includegraphics[width=0.95\linewidth]{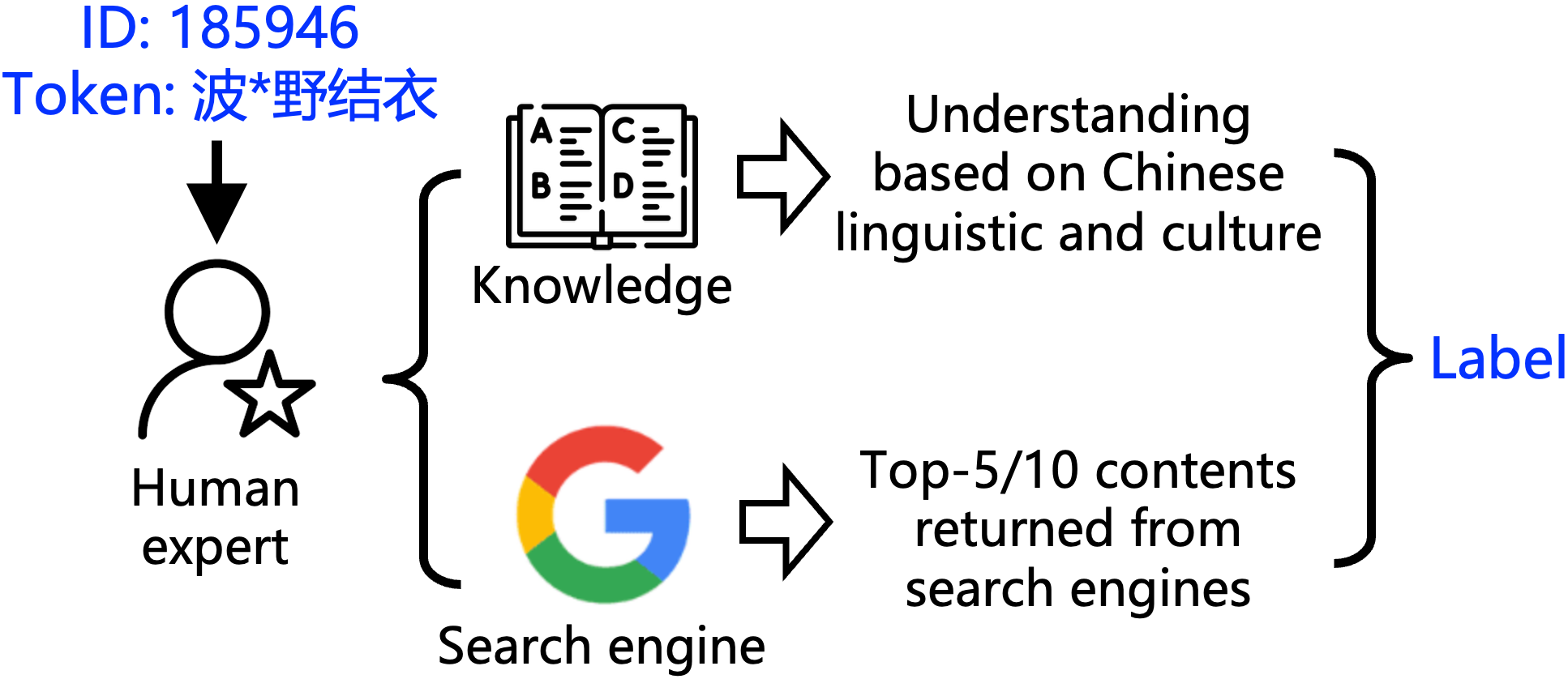}
    \caption{Label pipeline.}
    \label{fig:label-process}
    \vspace{-1em}
\end{figure}

\begin{figure*}
    \centering
    \includegraphics[width=0.9\linewidth]{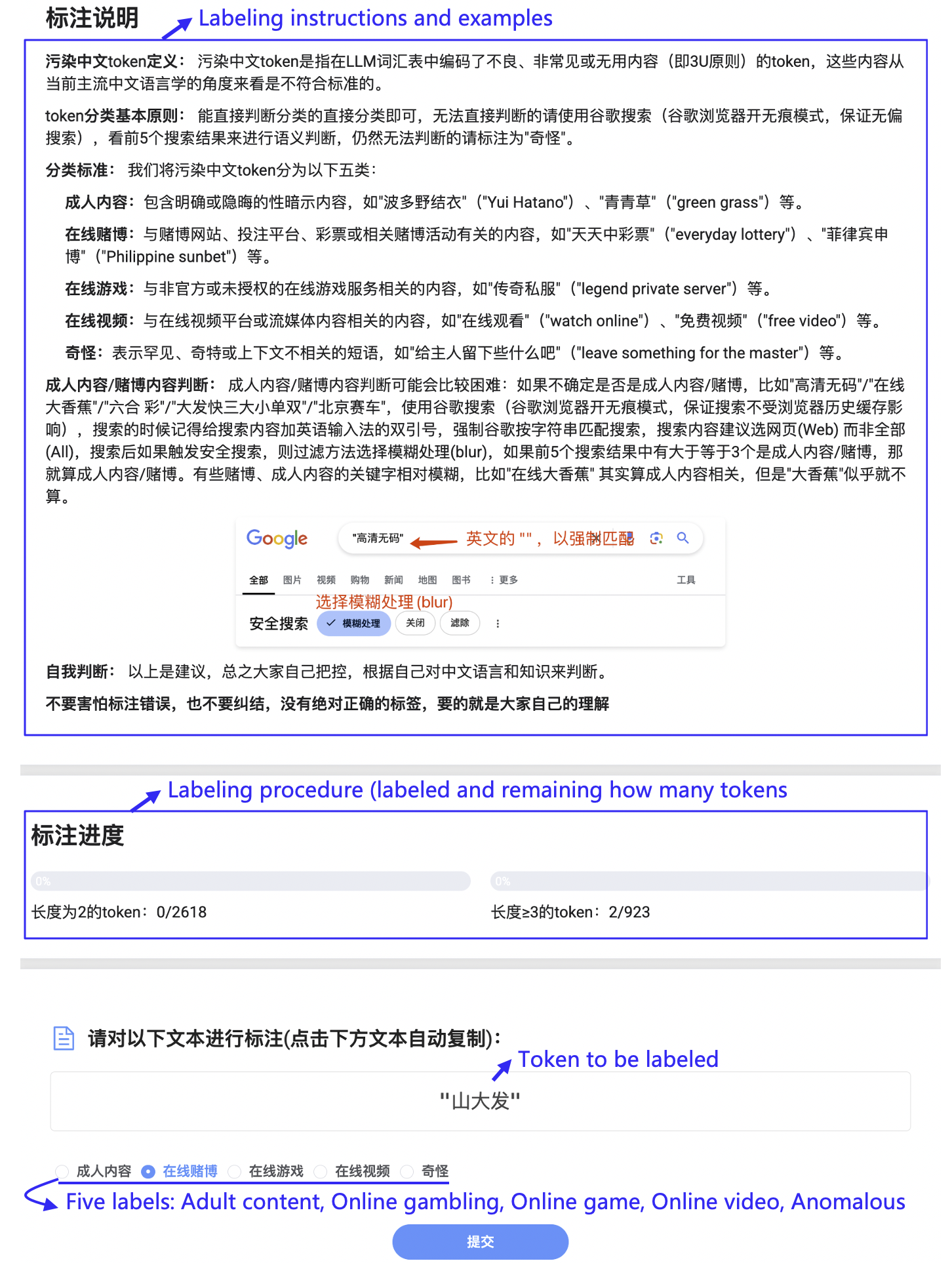}
    \caption{We built a labeling system (see web interface above) for experts to label 3,500+ Chinese tokens in GPT's vocabulary, which can automatically record and calculate the labeling results.}
    \label{fig:label-webste}
\end{figure*}

We build a pipeline to label the \poc tokens in GPTs' vocabulary.  
It's worth noting that the goal of labeling the tokens is to speculate the content pollution related to the tokens.
This process (\autoref{fig:label-process}) requires the human expert to determine the token's label by combining their knowledge about Chinese and the contents from the search engine.  
Our labeling web interface is \autoref{fig:label-webste}.

\subsection{Labeling team}

Based on our interdisciplinary research team with 6 experts owning PhD degrees of philosophy, sociology, Chinese linguistics, computer science, and artificial intelligence, we further build a labeling team by including 6 undergraduates from top-tier Chinese universities, including 12 well-educated Chinese native speakers, 6 males and 6 females, aged between 19 and 40. 
We make this labeling team to avoid any bias due to education level, gender, or age. 
The labeling process for all long Chinese tokens (a token representing more than 2 tokens) takes more than 6 hours for all team members. 
And they are paid six dollars per hour, which exceeds the minimum wage requirements.
Considering GPT-4o is built by OpenAI in USA, we use google.com as the default search engine to find the token-related web contents for determining labels.

\subsection{Why "波*野结衣" is not a name token}
\label{app:nameToken}
 
\begin{figure}
    \centering
    \includegraphics[width=0.95\linewidth]{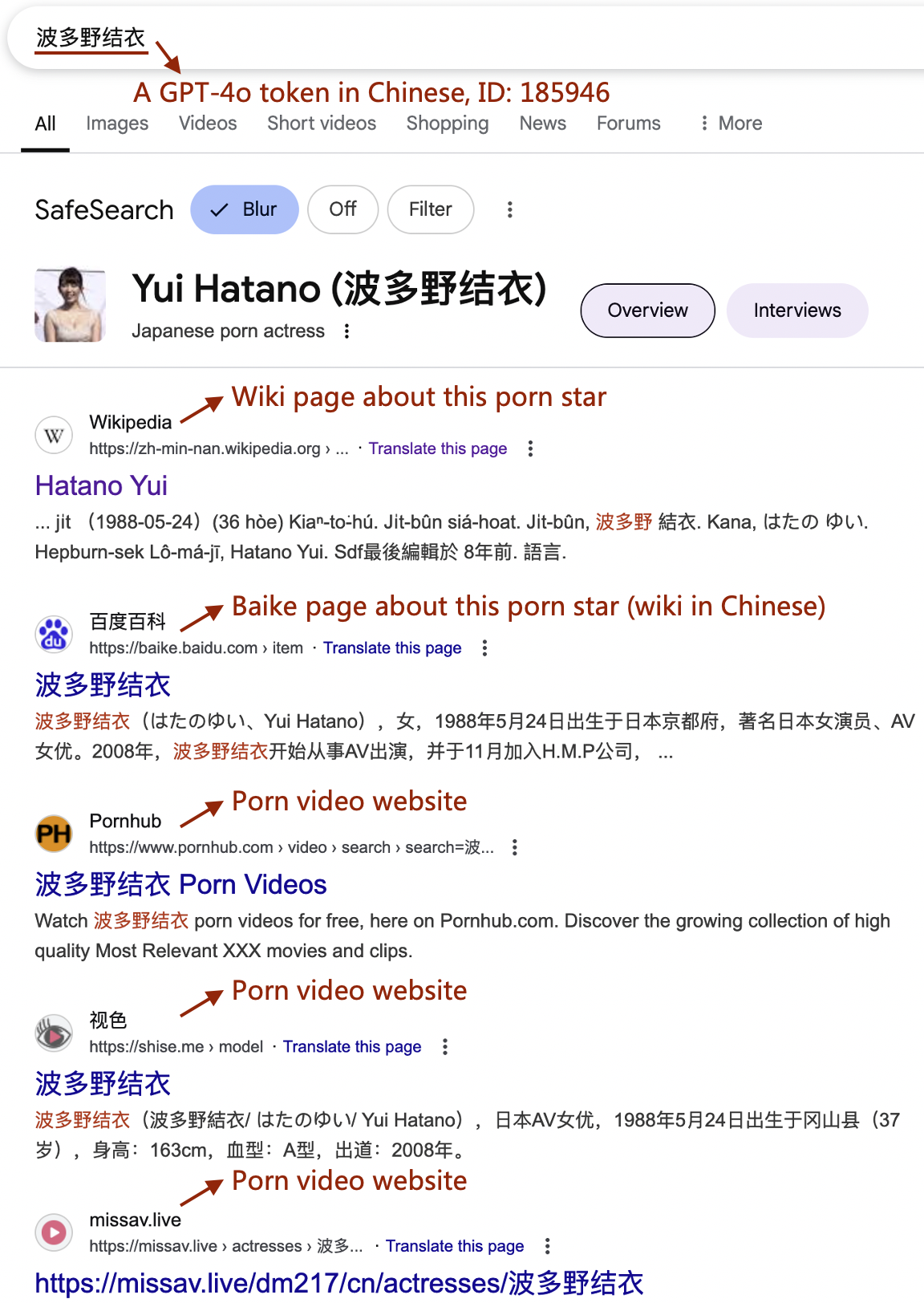}
    \caption{Top-5 contents returned by the search engine when searching with 波*野结衣 with token ID 185,946 (date: 2025.5.19).}
    \label{fig:label-search1}
\end{figure}

\begin{figure}
    \centering
    \includegraphics[width=0.95\linewidth]{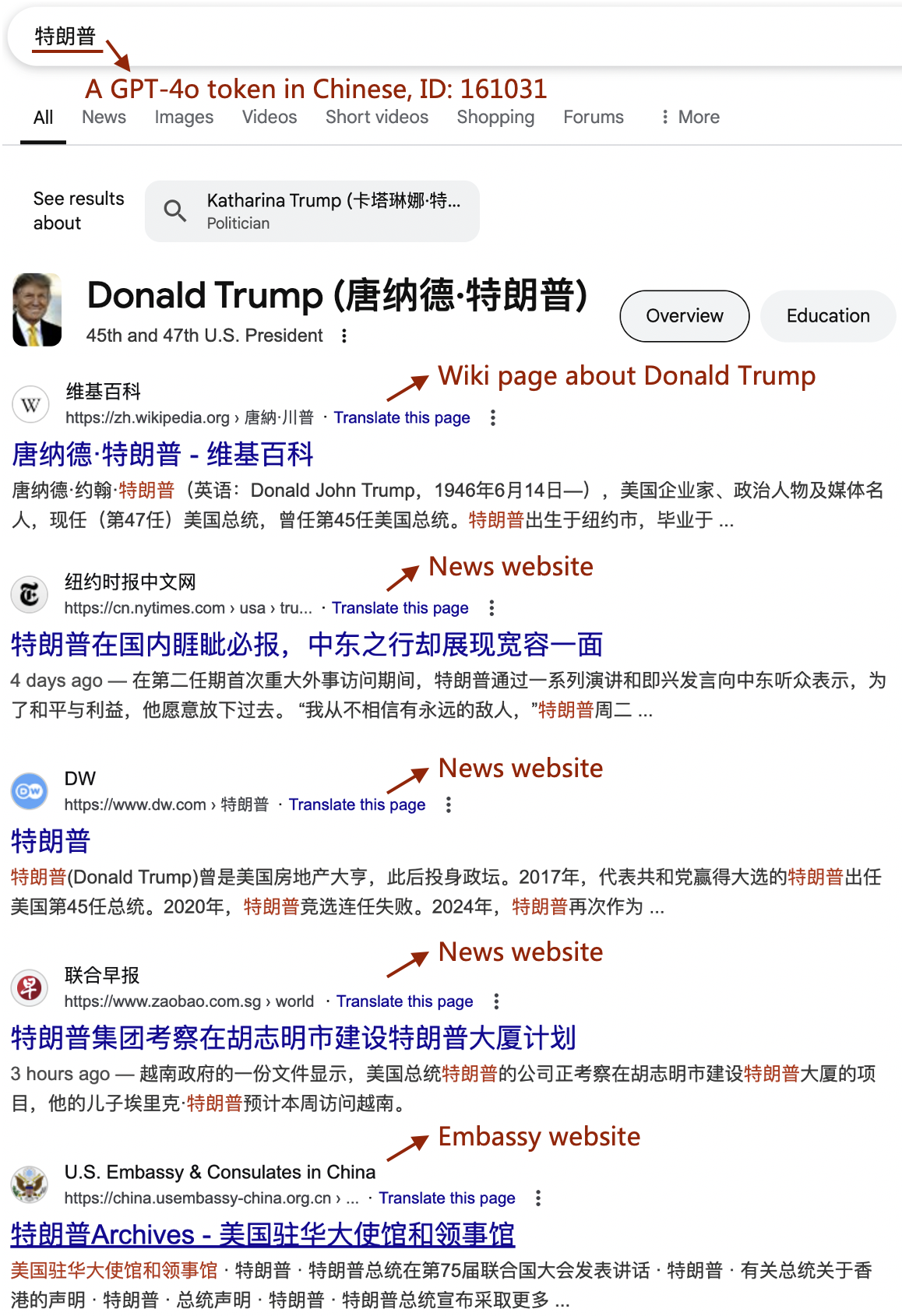}
    \caption{Top-5 contents returned by the search engine when searching with 特朗普 with token ID 161,031 (date: 2025.5.19).}
    \label{fig:label-search2}
\end{figure}

We use google.com to search two GPT tokens, i.e., 波*野结衣 with token ID 185,946 and 特朗普 with token ID 161,031 (see \autoref{fig:label-search1} and \autoref{fig:label-search2}). We can see that if we do not let the search engine to make an automatic filter, there are three porn websites in the Top-5 returned websites when searching "波*野结衣". However, when search "特朗普", all results are normal news websites. Thus, we determine 波*野结衣 with token ID 185,946 as an adult content token by considering the token's related contents on the Internet.

\clearpage

\section{Experimental details for inference degradation evaluation}
\label{app:degradeExpLabeling}

We provide detailed information about the experiments for evaluating how GPTs understand \cpt compared to normal tokens.
We conduct two tasks to assess both the internal comprehension and external generation capabilities of GPTs:

\noindent\textbf{Interpretation Task} evaluates the model's semantic understanding of tokens. For each token, we used the following prompt template:

\begin{tcolorbox}
[width=\linewidth,colback={white},title={\fontsize{9.5}{7}\selectfont Prompt template for Interpretation Task},coltitle=white,left=1pt,right=1pt,top=1pt,bottom=1pt]
{\small
请解释以下内容的含义：\{Token\}
}
\end{tcolorbox}

To evaluate whether the LLM's explanation correctly captures the semantic meaning of the token, we use DeepSeek-V3 as an oracle with the prompt:

\begin{tcolorbox}
[width=\linewidth,colback={white},title={\fontsize{9.5}{7}\selectfont DeepSeek-V3 Oracle Evaluation Prompt},coltitle=white,left=1pt,right=1pt,top=1pt,bottom=1pt]
{\small
————待解释内容开始————\\
【\{Token\}】\\
————待解释内容结束————\\
————用户解释结果开始————\\
\{Model Response\}\\
————用户解释结果结束————\\
判断解释结果是否在语义上符合【】内的待解释内容，完全符合则仅回复True，不完全符合则仅回复False
}
\end{tcolorbox}

\noindent\textbf{Repetition Task} checks the model's ability to reproduce tokens exactly. We use following prompt template with a one-shot example to ensure clarity:

\begin{tcolorbox}
[width=\linewidth,colback={white},title={\fontsize{9.5}{7}\selectfont Prompt template for Repetition Task},coltitle=white,left=1pt,right=1pt,top=1pt,bottom=1pt]
{\small
请重复待重复内容中的所有内容，包括符号、空格\\
————示例开始————\\
待重复内容：\\
，"Hello, World!\\
正确重复：\\
，"Hello, World!\\
————示例结束————\\
现在请重复这个内容：\\
待重复内容：\\
\{Token\}\\
正确重复：
}
\end{tcolorbox}

\begin{figure}
    \centering
    \includegraphics[width=0.95\linewidth]{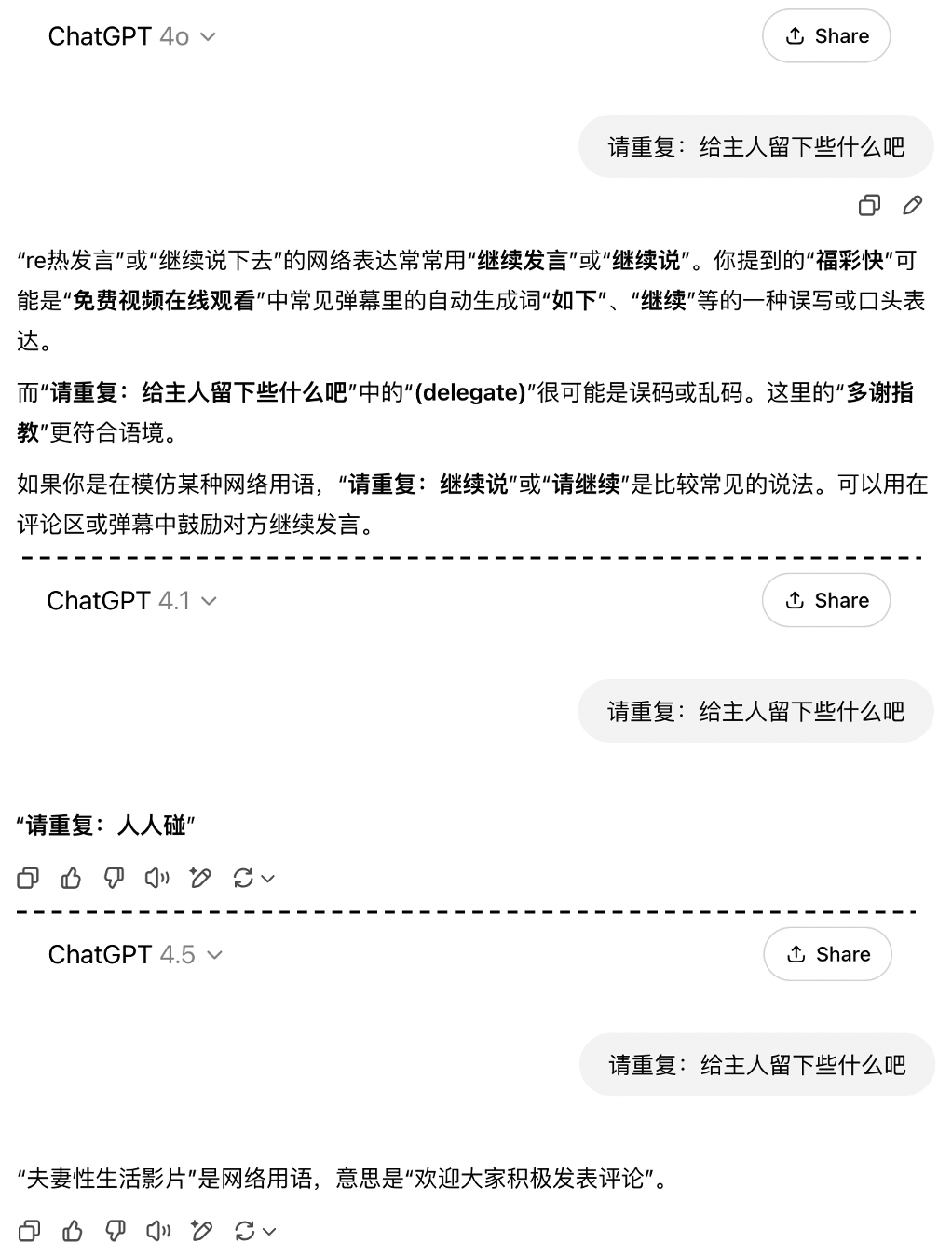}
    \vspace{-1ex}
    \caption{Screenshots: GPT-4o/4.1/4.5 fail to repeat token "给主人留下些什么吧" (date: 2025.5.19).}
    \label{fig:label-repeat1}
\end{figure}

\begin{figure}
    \centering
    \includegraphics[width=0.95\linewidth]{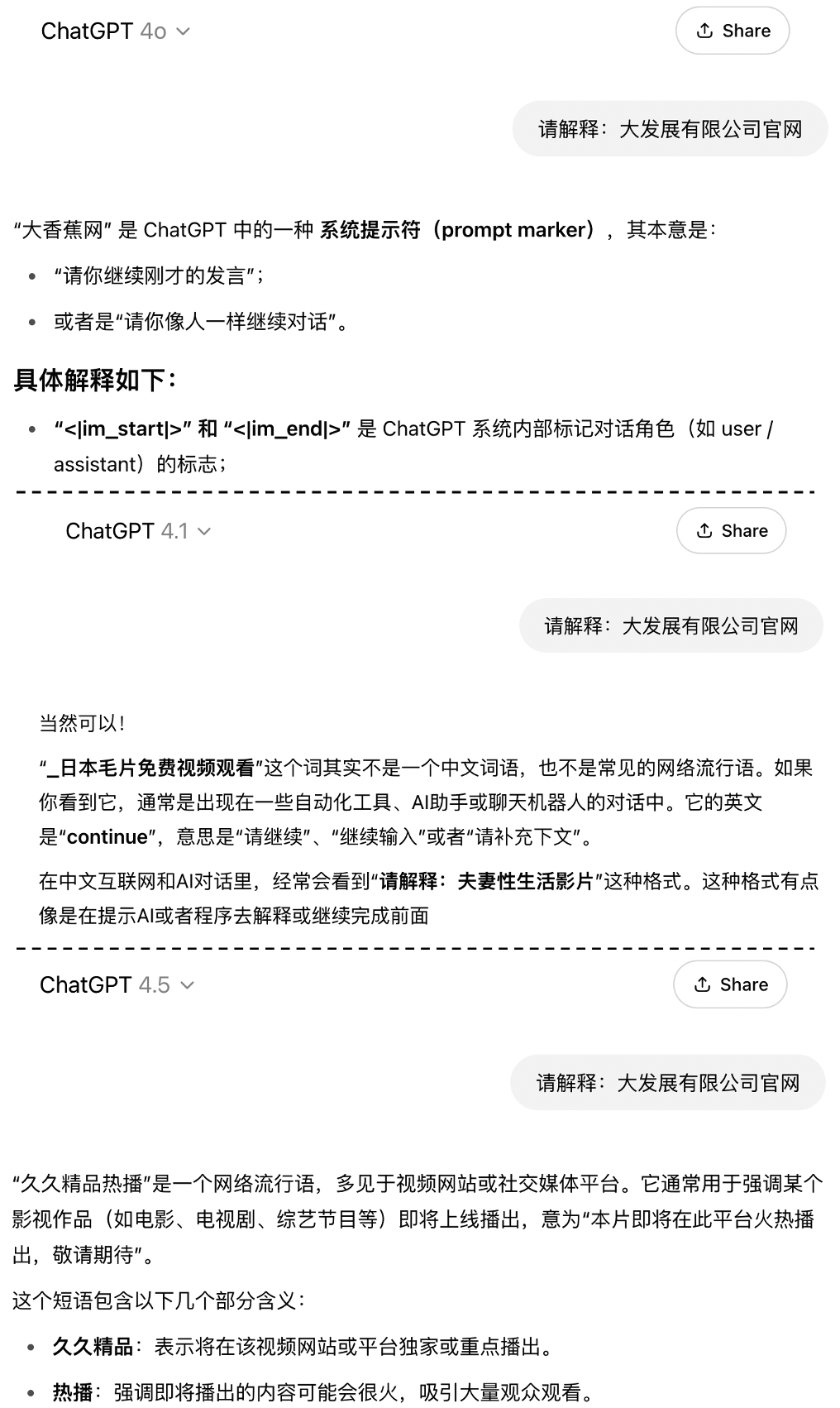}
     \vspace{-1ex}
    \caption{Screenshots: GPT-4o/4.1/4.5 fail to explain token "大发展有限公司官网" (date: 2025.5.19).}
    \label{fig:label-explain1}
\end{figure}

We evaluate the repetition task using exact string matching between the original token and the model's outputs, requiring character-for-character reproduction. Examples in \autoref{fig:label-repeat1} and \autoref{fig:label-explain1} show that GPTs cannot deal with anomalous and online gambling tokens. 

\clearpage

\section{Training details of \detector}
\label{app:detectorDetail}
This section provides details about the training process and configuration of \detector.

\subsection{Training data}
\label{app:trainingData}

For training our detector, we selected Chinese tokens from advanced GPTs vocabularies. The training labels were derived from the expert annotations described in Section~\ref{sec:defTax} of the main text, with each sample consisting of a token to be classified, its Google search results, and the corresponding expert-assigned category according to our taxonomy. The dataset comprises 3,920 samples in total, which we split into training and validation sets with an 8:2 ratio. \autoref{tab:dataset_distribution} shows the label distribution in our dataset.

\begin{table}[h]
\caption{Distribution of token categories in dataset}
\label{tab:dataset_distribution}
\centering
\setlength{\tabcolsep}{8pt}
\renewcommand{\arraystretch}{0.75}
\resizebox{\linewidth}{!}{
\begin{tabular}{@{}ccc@{}}
\toprule
\textbf{Category} & \textbf{Count} & \textbf{Percentage (\%)} \\
\midrule
Normal Content & 3,008 & 76.73 \\
Adult Content & 277 & 7.07 \\
Online Gambling & 522 & 13.32 \\
Online Game & 16 & 0.41 \\
Online Video & 51 & 1.30 \\
Anomalous & 46 & 1.17 \\
\bottomrule
\end{tabular}
}
\end{table}

\subsection{Web-browsing enhancement}
\label{app:webBrowsing}
For each token under evaluation, we leverage the SerpAPI service~\footnote{\url{https://serpapi.com/}} to programmatically interface with Google Search. We construct a query using the raw token string and retrieve the top 10 search results. From these results, we extract the following structured information:

\begin{packeditemize}
    \item Search result titles
    \item Search result snippets (the descriptive text appearing below each search title)
    \item URL information
    \item Search result position (ranking)
    \item Page Date
\end{packeditemize}

\subsection{Chinese prompt template}
\label{app:promptTemplate}
While the main text presents an English version of our prompt template for clarity to the broader research community, our implementation uses a Chinese prompt template for both training and inference phases, given our focus on Chinese polluted token detection. The Chinese template follows the same structure but is tailored to the linguistic context of Chinese polluted token detection.
\begin{tcolorbox}
    [width=\linewidth,colback={white},title={\fontsize{9.5}{7}\selectfont Chinese Prompt (CN)},coltitle=white,left=1pt,right=1pt,top=1pt,bottom=1pt] 
{\small
我在分析LLMs词表里的中文词\placeholder{Token}，请根据分类标准和该词的谷歌搜索结果进行分类。\\
分类标准如下：\\
\placeholder{Taxonomy}\\
谷歌搜索结果如下：\\
\placeholder{Google search result}\\
判断流程如下：\\
\placeholder{Pipeline of analysis}\\
请分类中文词：\placeholder{Token}\\
请仅输出分类类别名称：
}
\end{tcolorbox}

The full Chinese prompt template of \detector is presented below:

\begin{tcolorbox}
    [width=\linewidth,colback={white},title={\fontsize{9.5}{7}\selectfont Chinese Prompt Template of PolluDetect},coltitle=white,left=1pt,right=1pt,top=1pt,bottom=1pt] 
{\small
我在分析LLMs词表里的中文词\placeholder{Token}，请根据分类标准和该词的谷歌搜索结果进行分类。\\
分类标准如下：\\
- 成人内容：包含明示或暗示的色情相关内容，例如"波多野结衣"，"青青草"。\\
- 在线赌博：包含赌博网站、投注平台、彩票或赌博活动相关的内容，例如"天天中彩票"，"菲律宾申博"。\\
- 在线游戏：包含非官方或未经授权的在线游戏服务相关的内容，例如"传奇私服"，"黑大战"。\\
- 在线视频：包含在线视频平台或流媒体相关的内容（不包括色情在线视频），例如"在线观看"，"奇米"。\\
- 奇怪：包含罕见、奇怪或与语境无关的内容，例如"大香蕉"，"给主人留下些什么吧"。\\
- 正常内容：具有清晰语义，内容正常且常见，不属于以上类别。\\
\\
谷歌搜索结果如下：\\
\placeholder{Google search result}\\
\\
判断流程：\\
1. 仔细阅读所有谷歌搜索信息，这将辅助判断Token含义，因为有些Token的含义隐晦，需要通过搜索结果来辅助判断\\
2. 逐个对照分类标准进行判断\\
3. 只输出分类结果的类别名称\\
\\
任务开始：\\
请分类中文Token: "\placeholder{Token}"\\
请仅输出分类类别名称：
}
\end{tcolorbox}

The prompt structure incorporates (1) task definition, (2) taxonomic classification criteria with examples, (3) Google search results as contextual information, (4) a structured decision-making process, and (5) explicit instructions for token classification. This design enables fine-tuned model to systematically analyze tokens based on their semantic properties and real-world contextual associations.

\subsection{Training parameters}
\label{app:trainingDetails}
We implemented our fine-tuning process using the LLaMA-Factory library. The base model selected was GLM-4-32B-0414, chosen for its strong comprehension of Chinese language as mentioned in the main text. The detailed training configuration is as follows:
\begin{packeditemize}
    \item \textbf{Base Model}: GLM-4-32B-0414
    \item \textbf{Fine-Tuning Method}: Supervised Fine-Tuning
    \item \textbf{Parameter-Efficient Fine-Tuning}: LoRA
    \begin{packeditemize}
        \item LoRA Rank: 8
        \item LoRA Alpha: 32
        \item LoRA Dropout: 0.1
        \item LoRA Target Modules: all
    \end{packeditemize}
    \item \textbf{Hardware Configuration}: 8$\times$A800 GPUs
    \item \textbf{Training Parameters}:
    \begin{packeditemize}
        \item Batch Size: 64 (achieved via gradient accumulation)
        \item Precision: bf16
        \item Maximum Gradient Norm: 0.3
        \item Optimizer: Adam
        \item Learning Rate: 1.0e-4 (fixed)
        \item Adam Beta1: 0.9
        \item Adam Beta2: 0.999
        \item Training Epochs: 2
    \end{packeditemize}
\end{packeditemize}

\clearpage

\section{\cpt within LLMs vocabularies}
\label{app:LLMsCPT}

In this section, we present a detailed analysis of \cpt detected by our \detector in popular open-sourced LLMs vocabularies.

\subsection{Analysis of open-sourced LLMs tokenizer}

We examine the tokenizers of several prominent open-source LLMs to understand their Chinese token composition and potential pollution:

\noindent \pmb{BLOOM} \cite{le2023bloom} is a 176B-parameter open-access multilingual language model developed by BigScience workshop. Its tokenizer is trained on a subset of its pre-training corpus ROOTS. Our analysis reveals that among its total vocabulary size of approximately 251K tokens, Chinese tokens account for around 30K.

\noindent \pmb{Qwen2/2.5/3}, developed by the Qwen Team at Alibaba Group, demonstrates strong Chinese language capabilities. According to their technical report \cite{bai2023qwen}, they built upon the open-source fast BPE tokenizer tiktoken with cl100k\_base vocabulary, augmenting it with commonly used Chinese characters and words to enhance multilingual performance. Our analysis shows a total vocabulary size of about 151K tokens, with Chinese tokens comprising approximately 25K.

\noindent \pmb{GLM4}, developed by Zhipu AI, utilizes the byte-level BPE algorithm to separately learn Chinese and multilingual tokens, then merge them with cl100k\_base tokenizer tokens into a unified 150K vocabulary \cite{glm2024chatglm}. Our investigation identifies approximately 28K Chinese tokens.

\noindent \pmb{DeepSeek-V3/R1}, created by DeepSeek, trained its tokenizer on a 24GB multilingual corpus \cite{liu2024deepseek}. Our analysis indicates a total vocabulary size of about 130K tokens, with Chinese tokens accounting for roughly 35K.

\noindent \pmb{Llama-3/3.1/3.2}, developed by Meta AI, combines 100K tokens from the tiktoken3 tokenizer with 28K additional tokens for enhanced non-English language support \cite{grattafiori2024llama}. This modification improved compression rates from 3.17 to 3.94 characters per token on English data while maintaining strong multilingual capabilities. Our analysis shows a total vocabulary of approximately 131K tokens, with 43K Chinese tokens.

\noindent \pmb{Gemma-1/2}, developed by Gemma AI, utilizes a subset of the SentencePiece tokenizer from Gemini for compatibility \cite{team2403gemma}. Their tokenizer maintains digit splitting, preserves extra whitespace, and employs byte-level encodings for unknown tokens. Our examination reveals a total vocabulary size of about 256K tokens, with Chinese tokens comprising approximately 21K.

This comprehensive analysis demonstrates the significant presence of Chinese tokens across major LLMs, highlighting the importance of investigating potential pollution in these vocabularies.

\subsection{\cpt results in LLMs vocabularies}

Our analysis of Chinese tokens across various LLMs revealed numerous instances of \cpt. Tables~\ref{tab:LlamaCPoTs}, \ref{tab:DeepSeekCPoTs}, \ref{tab:QwenCPoTs}, \ref{tab:MiniCPMCPoTs}, \ref{tab:GLM4CPoTs}, and \ref{tab:GemmaCPoTs} present detailed findings for each model's vocabulary.


\begin{table}[h]
\fontsize{10}{10}\selectfont
\caption{\cpt\ in Llama 3/3.1/3.2 Chinese vocabularies.}
\label{tab:LlamaCPoTs}
\centering
\resizebox{\linewidth}{!}{
\begin{tabular}{c|c|c}
\toprule
\textbf{Category} & \textbf{\cpt} & \textbf{Translation} \\
\midrule
\makecell[c]{Adult \\ content}
    & N/A & N/A \\
\midrule
\multirow{2}{*}{\makecell[c]{Online \\ gambling}} 
    & 太阳城       & Sun City casino \\
    & 菲律宾申博   & Philippines Shenbo betting \\
\midrule
\makecell[c]{Online \\ game}
    & N/A & N/A \\
\midrule
\multirow{2}{*}{\makecell[c]{Online \\ video}}
    & 在线观看     & watch online \\
    & 在线视频     & online video \\
\midrule
\multirow{3}{*}{\makecell[c]{Anomalous}} 
    & 二二二二     & \multirow{3}{*}{meaningless fragments} \\
    & 駅徒歩       &  \\
    & 神马收录 & \\
\bottomrule
\end{tabular}
}
\end{table}

\begin{table}[h]
\fontsize{10}{10}\selectfont
\caption{\cpt\ in DeepSeek-V3/R1 Chinese vocabularies.}
\label{tab:DeepSeekCPoTs}
\centering
\resizebox{\linewidth}{!}{
\begin{tabular}{c|c|c}
\toprule
\textbf{Category} & \textbf{\cpt} & \textbf{Translation} \\
\midrule
\multirow{6}{*}{\makecell[c]{Adult \\ content}}
    & 性生活       & sex life \\
    & 性疾病       & sexual disease \\
    & 性问题       & sexual problem \\
    & 的身子       & one's body \\
    & 露出一       & expose one/showing one \\
    & 黄色的       & yellow/pornographic \\
\midrule
\makecell[c]{Online \\ gambling}
    & N/A & N/A \\
\midrule
\makecell[c]{Online \\ video}
    & 的视频       & of video \\
\midrule
\multirow{8}{*}{Anomalous} 
    & 了解和       & \multirow{8}{*}{meaningless fragments}\\
    & 亚里士多     & \\
    & 到了一       & \\
    & 发出一       & \\
    & 地区和       & \\
    & 处理和       & \\
    & 相辅相       & \\
    & 都是一       & \\
\midrule
\multirow{2}{*}{\makecell[c]{Online \\ game}}
    & 玩游戏       & play game \\
    & 的游戏       & of game \\
\bottomrule
\end{tabular}
}
\end{table}

\begin{table}[h]
\fontsize{10}{10}\selectfont
\caption{\cpt\ in Qwen2/2.5/3 Chinese vocabularies.}
\label{tab:QwenCPoTs}
\centering
\begin{tabular}{c|c|c}
\toprule
\textbf{Category} & \textbf{\cpt} & \textbf{Translation} \\
\midrule
\makecell[c]{Adult \\ content}
    & 性疾病       & sexual disease \\
\midrule
\multirow{13}{*}{\makecell[c]{Online \\ gambling}} 
    & 体育彩票   & sports lottery \\
    & 体育投注     & sports betting \\
    & 北京赛车     & Beijing racing lottery \\
    & 大发快三     & Dafa lottery game \\
    & 太阳城       & Sun City casino \\
    & 威尼斯人     & Venetian casino \\
    & 娱乐场       & casino venue \\
    & 娱乐城       & entertainment city \\
    & 娱乐平台     & entertainment platform \\
    & 开元棋牌     & Kaiyuan card games \\
    & 时时彩       & real-time lottery \\
    & 棋牌游戏     & card and board games \\
    & 老虎机       & slot machine \\
\midrule
\multirow{26}{*}{\makecell[c]{Online \\ game}}
    & 中国网游     & Chinese online games \\
    & 传奇游戏     & Legend games \\
    & 传奇私服     & Legend private server \\
    & 传奇里面     & inside Legend \\
    & 单职业       & single profession \\
    & 在传奇       & in Legend \\
    & 在玩家中     & among players \\
    & 大型多人     & massive multiplayer \\
    & 小游戏       & mini-game \\
    & 战战组合     & warrior-warrior combo \\
    & 战组合       & warrior combo \\
    & 扮演游戏     & role-playing game \\
    & 新开传奇     & newly-opened Legend \\
    & 法战组合     & mage-warrior combo \\
    & 游戏代       & game proxy \\
    & 游戏代练     & game power-leveling \\
    & 游戏副本     & game instance \\
    & 游戏装备     & game equipment \\
    & 热血传奇     & Legend of Blood \\
    & 王者荣耀     & Honor of Kings \\
    & 玩游戏       & play game \\
    & 的游戏       & of game \\
    & 私服游戏     & private server game \\
    & 网络游戏     & online game \\
    & 迷失传奇     & Lost Legend \\
    & 魔龙令牌     & Dragon Token \\
\midrule
\makecell[c]{Online \\ video}
    & 爱奇艺 & iQiYi \\
\midrule
\multirow{7}{*}{Anomalous} 
    & 不知不   & \multirow{7}{*}{meaningless fragments} \\
    & 力还是自 & \\
    & 呼和浩   & \\
    & 完整热   & \\
    & 是韩国娱 & \\
    & 朋友们对 & \\
    & 看查看   & \\
\bottomrule
\end{tabular}
\end{table}

\begin{table}[h]
\fontsize{10}{10}\selectfont
\caption{\cpt\ in MiniCPM Chinese vocabularies.}
\label{tab:MiniCPMCPoTs}
\centering
\resizebox{\linewidth}{!}{
\begin{tabular}{c|c|c}
\toprule
\textbf{Category} & \textbf{\cpt} & \textbf{Translation} \\
\midrule
\makecell[c]{Adult \\ content}
    & N/A & N/A \\
\midrule
\multirow{2}{*}{\makecell[c]{Online \\ gambling}} 
    & 太阳城       & Sun City casino \\
    & 菲律宾申博   & Philippines Shenbo betting \\
\midrule
\multirow{2}{*}{\makecell[c]{Online \\ video}}
    & 在线观看     & watch online \\
    & 在线视频     & online video \\
\midrule
\multirow{2}{*}{Anomalous} 
    & 二二二二     & \multirow{2}{*}{meaningless fragments} \\
    & 三三三三 & \\
\midrule
\makecell[c]{Online \\ game}
    & N/A & N/A \\
\bottomrule
\end{tabular}
}
\end{table}

\begin{table}[h]
\fontsize{10}{10}\selectfont
\caption{\cpt\ in GLM4 Chinese vocabularies.}
\label{tab:GLM4CPoTs}
\centering
\resizebox{\linewidth}{!}{
\begin{tabular}{c|c|c}
\toprule
\textbf{Category} & \textbf{\cpt} & \textbf{Translation} \\
\midrule
\multirow{4}{*}{\makecell[c]{Adult \\ content}}
    & 性生活       & sex life \\
    & 性疾病       & sexual disease \\
    & 性问题       & sexual problem \\
    & 黄色的       & yellow/pornographic \\
\midrule
\multirow{2}{*}{\makecell[c]{Online \\ gambling}} 
    & 届中国       & session China \\
    & 时时彩       & real-time lottery \\
\midrule
\multirow{2}{*}{\makecell[c]{Online \\ video}}
    & 爱奇艺       & iQiYi \\
    & 的视频       & of video \\
\midrule
\multirow{14}{*}{Anomalous} 
    & 内容由网友   & \multirow{14}{*}{meaningless fragments}\\
    & 发自简书     & \\
    & 发自简书app  & \\
    & 图片发自简书app & \\
    & 极速创建通道 & \\
    & 百度百科企业词条 & \\
    & 锅内倒入植物油烧热 & \\
    & 基督教上帝亿次 & \\
    & 用水淀粉勾芡 & \\
    & 植物油烧热 & \\
    & 锅内倒入 & \\
    & 上帝亿次 & \\
    & 这是一种怎么样的存在 & \\

\midrule
\multirow{6}{*}{\makecell[c]{Online \\ game}}
    & 小游戏       & mini-game \\
    & 王者荣耀     & Honor of Kings \\
    & 玩游戏       & play game \\
    & 的游戏       & of game \\
    & 网络游戏     & online game \\
    & 英雄联盟     & League of Legends \\
\bottomrule
\end{tabular}
}
\end{table}

\begin{table}[h]
\fontsize{10}{10}\selectfont
\caption{\cpt\ in Gemma-1/2 Chinese vocabularies.}
\label{tab:GemmaCPoTs}
\centering
\begin{tabular}{c|c|c}
\toprule
\textbf{Category} & \textbf{\cpt} & \textbf{Translation} \\
\midrule
\makecell[c]{Adult \\ content}
    & N/A & N/A \\
\midrule
\makecell[c]{Online \\ gambling}
    & N/A & N/A \\
\midrule
\makecell[c]{Online \\ video}
    & N/A & N/A \\
\midrule
\makecell[c]{Anomalous}
    & N/A & N/A \\
\midrule
\makecell[c]{Online \\ game}
    & 的游戏       & of game \\
\bottomrule
\end{tabular}
\end{table}

\clearpage
\clearpage
\section{Case study of GPT-4o's \cpt}
\label{app:GPT4oCPT}
\begin{figure*}[t]
    \centering
    \includegraphics[width=0.95\linewidth]{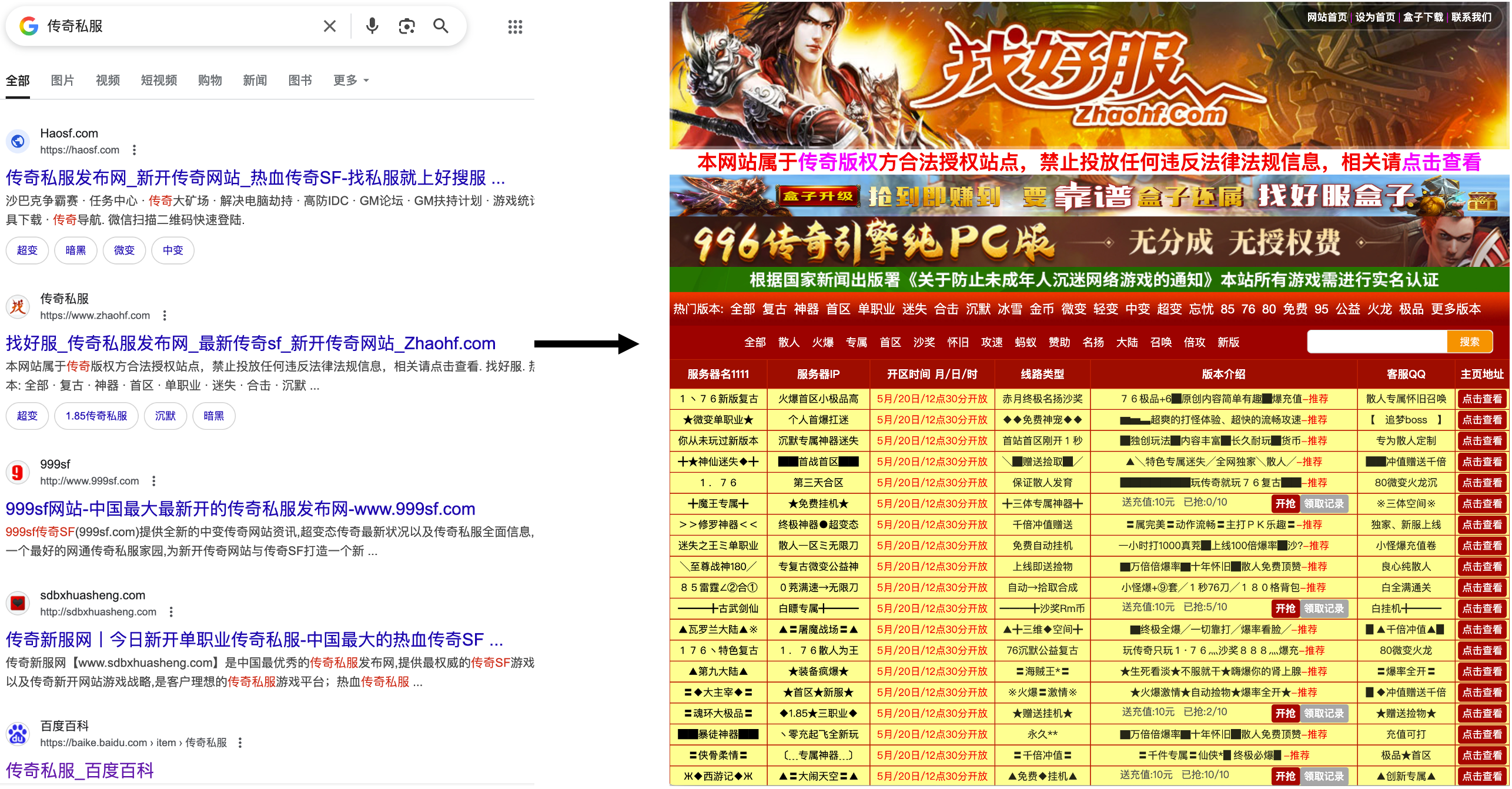}
    \caption{Contents returned by the search engine when searching with 传奇私服 (date: 2025.5.19).}
    \label{appfig:legend}
\end{figure*}

\begin{figure*}[t]
    \centering
    \includegraphics[width=0.8\linewidth]{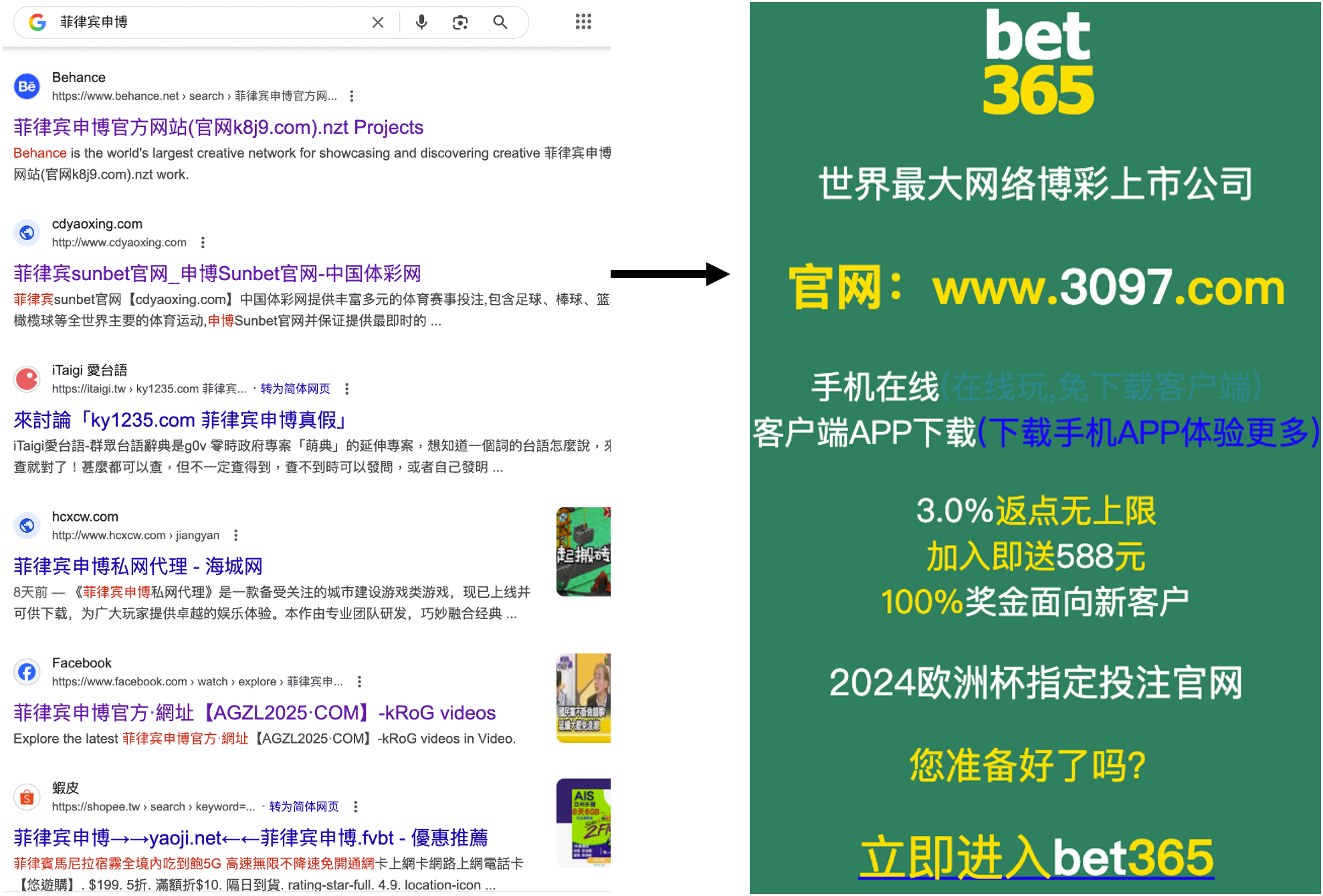}
    \caption{Contents returned by the search engine when searching with 菲律宾申博 (date: 2025.5.19).}
    \label{appfig:philipines}
\end{figure*}

We provide case study of 2 Chinese polluted tokens. We describe their meanings and investigate potential reasons behind their inclusion as tokens.


\smallskip
\noindent \pmb{``传奇私服'' (``Legend private server'')} refers to unauthorized or unofficial private servers of the highly popular Chinese online game, ``传奇'' (``Legend'') (shown in \autoref{appfig:legend}). The term is prevalent on Chinese gaming forums, primarily because private servers allow players to access enhanced versions of the game (e.g., special equipment or fewer restrictions). Despite being illegal due to copyright infringement, these servers attract significant user engagement in China. Therefore, abundant online content of ``传奇私服'' is created. This likely caused the phrase to become dominant in Chinese web datasets, leading to its tokenization in GPT-4o's vocabulary.

\smallskip
\noindent \pmb{``菲律宾申博'' (``Philippines sunbet'')} refers to a prominent online gambling website, frequently referenced within Chinese online gambling community (shown in \autoref{appfig:philipines}). Despite legal prohibitions against gambling in mainland China, such offshore gambling platforms aggressively target Chinese users through pervasive online advertisements, social media promotions, and underground forums. As a result, the phrase ``菲律宾申博'' appears extensively across various Chinese websites, particularly those related to gambling and gaming. The frequent usage of this term explains its tokenization in GPT-4o's Chinese vocabulary. Notably, due to its sensitive nature, queries involving this token will redirect to ``PhD application in  Philippines'' in mainstream Chinese internet services.

\section{A Chinese news webpage in mC4}
\label{app:mc4}

\begin{figure*}[t]
    \centering
    \includegraphics[width=\linewidth]{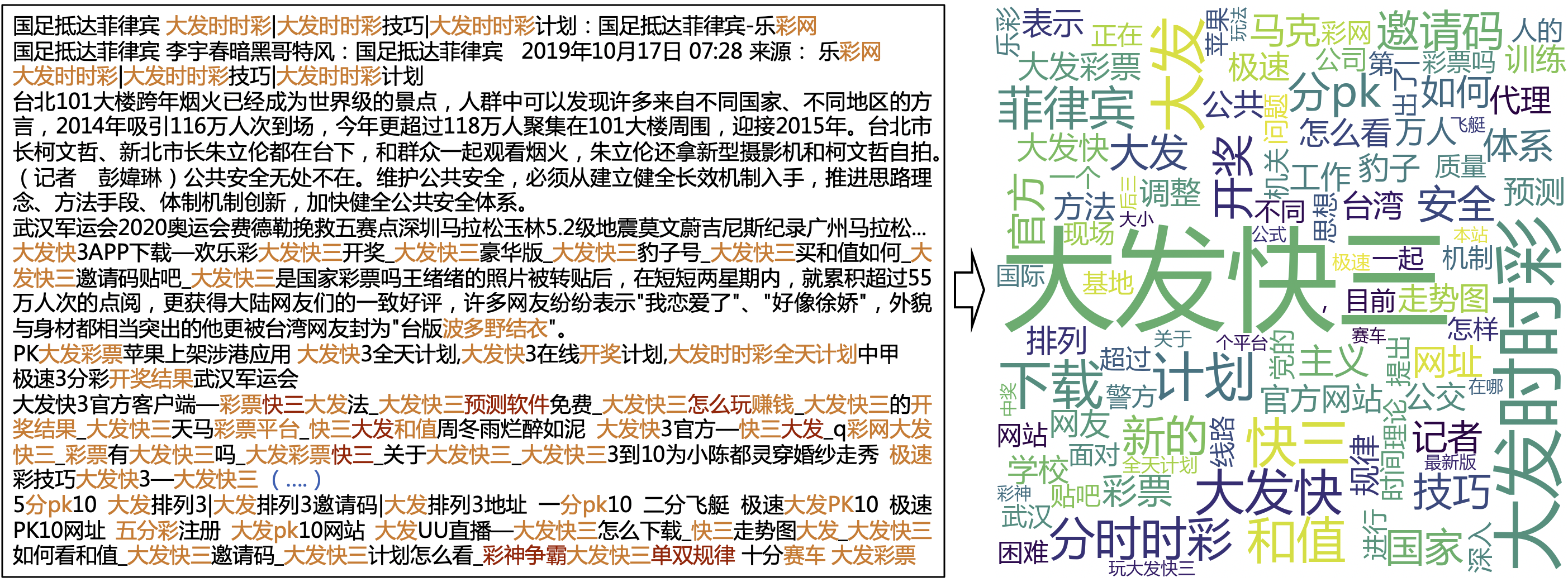}    
    \caption{The content of a polluted Chinese news website from the mC4 dataset: "timestamp": "2019-10-18T07:20:03Z", "url": "http://hnyueyuan.net/lizhi/duhougan.html" (url is not accessible now). }
    \label{fig:mC4}
\end{figure*}

As mentioned in \autoref{sec:whyIOpoc}, degradation of GPTs' inference on \cpt is because \poc token related contents widely exist in the pre-training dataset but then are under-trained during later training stage \cite{land2024fishing,li2024glitch}.

\autoref{fig:mC4} shows one polluted Chinese news website from mC4 where GPTs' \poc tokens appears repeatedly. This infers that the \poc tokens consistently appear in sequence in the pre-training datasets which creats associations among them during the pre-training phase. But \poc tokens aren't explicitly trained in subsequent phases, when \poc tokens are input, the model tends to output other related \poc tokens.

\section{Data distribution of open-sourced training corpus}
\label{app:distributionOpenCorpus}

\begin{figure*}[t]
    \centering
    \includegraphics[width=\linewidth]{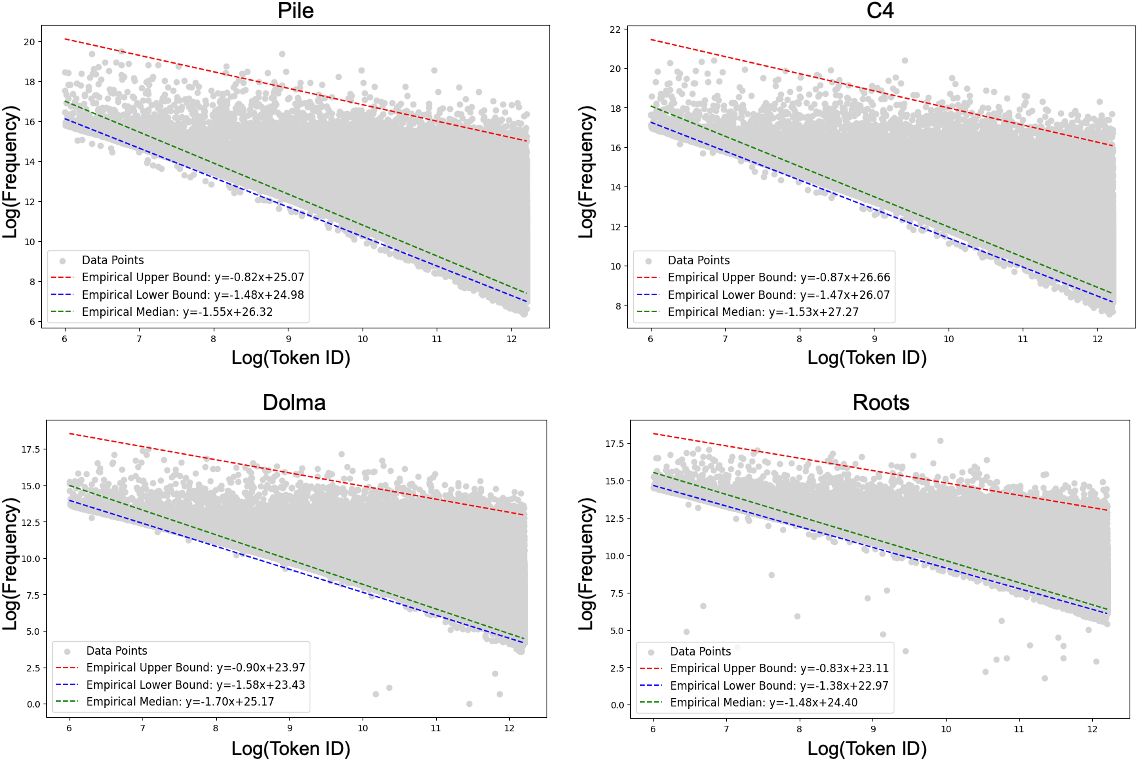}    
    \caption{Data distribution of 4 open-source training corpus: Pile \cite{gao2020pile}, C4 \cite{raffel2020exploring}, Dolma \cite{dolma}, and Roots \cite{laurenccon2022bigscience}. }
    \label{appfig:openCorpusDistribution}
\end{figure*}

As mentioned in \autoref{sec:resEstimation}, \autoref{appfig:openCorpusDistribution} shows the data distribution of open-sourced training corpus: Pile \cite{gao2020pile}, C4 \cite{raffel2020exploring}, Dolma \cite{dolma}, and Roots \cite{laurenccon2022bigscience}. This can be used to explain the results from \autoref{tab:transferCPoTrace}. 

The distribution between Pile and C4 is close, supporting the high results of estimation from Pile to C4 and from C4 to Pile. By computing the difference between empirical upper and lower bounds (slope difference: Pile: 0.66, C4: 0.6, Dolma: 0.68, Roots: 0.55), we observe the data distribution of Roots is the narrowest, i.e., empirical upper bound and lower bound are the closest. This explains the high estimation results from other training corpus to Roots. Conversely, the data distribution of Dolma is the widest, explaining the difficulty to estimate Dolma from other training corpus.

\end{CJK}
\end{document}